# Interpretable Clustering with Adaptive Heterogeneous Causal Structure Learning in Mixed Observational Data

Wenrui Li, Qinghao Zhang, and Xiaowo Wang*

***Abstract*—**Understanding causal heterogeneity is essential for scientific discovery in domains such as biology and medicine. However, existing methods lack causal awareness, with insufficient modeling of heterogeneity, confounding, and observational constraints, leading to poor interpretability and difficulty distinguishing true causal heterogeneity from spurious associations. We propose an unsupervised framework, **HCL** (Interpretable Causal Mechanism-Aware Clustering with Adaptive Heterogeneous Causal Structure Learning), that jointly infers latent clusters and their associated causal structures from mixed-type observational data without requiring temporal ordering, environment labels, interventions or other prior knowledge. HCL relaxes the homogeneity and sufficiency assumptions by introducing an equivalent representation that encodes both structural heterogeneity and confounding. It further develops a bi-directional iterative strategy to alternately refine causal clustering and structure learning, along with a self-supervised regularization that balance cross-cluster universality and specificity. Together, these components enable convergence toward interpretable, heterogeneous causal patterns. Theoretically, we show identifiability of heterogeneous causal structures under mild conditions. Empirically, HCL achieves superior performance in both clustering and structure learning tasks, and recovers biologically meaningful mechanisms in real-world single-cell perturbation data, demonstrating its utility for discovering interpretable, mechanism-level causal heterogeneity.

## I. INTRODUCTION

Understanding causal relationships lies at the core of scientific discovery, offering insights not only into what correlates, but why and how changes propagate across complex systems [1]. Unlike correlational models, causal models aim to uncover the underlying mechanisms that govern the systematic behavior and evolution [2],[3]. Accurate modeling of causal mechanisms enables reliable predictions, interpretable decision-making, and targeted intervention across domains such as biology [4], medicine [5], economics [6].

A fundamental task of causal modeling is causal structure learning (CSL), namely recovering a directed acyclic graph (DAG) that encodes a chain of causal relationships (causal structure) through which one variable affect others from observational data. In contrast to spurious correlations that often fail to generalize under distributional shifts, causal structures encode the underlying mechanism expected to remain invariant across varying environments [7],[8]. However, in many real-world scenarios—such as under drug perturbations—many underlying mechanisms may be subject to direct influence, necessitating the modeling of how causal mechanisms are modified by such environmental shifts. More generally, causal structures can change when environmental conditions shift significantly [9]. For instance, cells under different lineages, drug perturbations, temporal progressions, or disease states may express the same genes, yet exhibit distinct regulatory relationships among them, resulting in heterogeneous causal graphs across conditions. Neglecting such structural heterogeneity not only hinders mechanistic insight and generalization but also limits the potential for precise and context-aware interventions.

### *A. Prior Work*

Existing CSL methods are mainly grounded in the Structural Causal Model (SCM) framework [7],[8]. These approaches can be broadly categorized into four families: constraint-based, score-based, hybrid, and function-based methods [10]. Constraint-based methods [11],[12] rely on conditional independence (CI) tests to infer the equivalence class of DAGs. Score-based methods [13],[14] formulate CSL as a combinatorial optimization problem. They define a goodness-of-fit score to quantify the alignment between candidate DAGs and the data, and search for equivalence class of DAGs that maximizes this score. Hybrid methods [15] combine the constraint-based pruning with the score-based search but may inherit limitations from both paradigms. Function-based methods directly model the functional dependencies between variables. They often assume that each variable is generated as a deterministic function of its parents and an independent exogenous variable and infer the causal graph by leveraging asymmetries in functional form or distribution. Representative methods, such as NOTEARS [16] and KEEL [17], parameterize the DAG as a weighted adjacency matrix and enforce graph properties through smooth algebraic constraints, enabling structure learning through gradient descent.

Despite considerable progress, existing CSL methods have predominantly relied on the strict homogeneity assumption that all samples are generated from an identical causal graph. While effective in homogeneous settings, this assumption limits their applicability to real-world scenarios characterized by structural heterogeneity. Enforcing a one-size-fits-all model in such cases leads to spurious structures and misleading interpretations, as these methods are fundamentally incapable of capturing environment-specific causal variations.

In response to this limitation, several studies have attempted to identify causal structure variability by leveraging temporal information, environmental annotations or intervention targets [18-22]. However, these resources are rarely available in many

Wenrui Li is with the Department of Automation, Tsinghua University, Haidian, Beijing, China (e-mail: li-wr23@mails.tsinghua.edu.cn).

Qinghao Zhang is with the Department of Electrical Engineering, Tsinghua University, Haidian, Beijing, China.

Xiaowo Wang (*Corresponding Author*) is with the Department of Automation, Tsinghua University, Haidian, Beijing, China.

observational settings. Moreover, they often assume fixed causal skeleton or smooth, gradual changes in causal mechanisms, limiting their capacity to capture latent modifiers with environment-specific effects and abrupt, sparse structural variations. Furthermore, existing methods face a trade-off between heterogeneity and invariance. On one hand, some methods focus on shared, invariant structures but may overlook context-specific causal variations [23-26]. On the other hand, other approaches capture structural differences but may sacrifice model scalability, lack cross-domain transferability and suffer from overfitting to spurious noise [27],[28]. This dichotomy reflects a deeper challenge: generalization in heterogeneous systems requires balancing invariant causal regularities with context-specific variations. Neglecting either aspect limits the model's generalization.

The adaptive identification of heterogenous causal structures from observational data in an unsupervised manner remains an unsolved challenge. In heterogenous scenarios, the joint distribution no longer adheres to a single graph, as causal structures may vary across latent context, resulting in an exponential increase in the effective search space of graph. Recent attempts often cluster samples based on statistical similarity prior to CSL [27, 29-31]. While intuitive, such methods overlook unmeasured confounders and bias (e.g. noise, sample scarcity or selection bias). Clustering based on statistical similarity cannot reliably distinguish between true causal heterogeneity and spurious variation induced by confounding or bias. This conflation risks overestimating heterogeneity, producing uninterpretable clusters and spurious stratification that propagate errors into subsequent steps. Furthermore, their CSL steps rely on strong assumptions such as causal sufficiency and faithfulness [32],[33], implicitly presuming latent confounder-free and unbiased environments — conditions which rarely met in practice. Violations of these assumptions can introduce spurious associations that distort structure learning and ultimately undermine the discovery of heterogenous causal patterns.

This challenge stems not merely from empirical limitations or algorithmic complexity, but from the following fundamental barriers.

First, simultaneous estimation of heterogenous causal structures and sample clustering entails solving a high-dimensional, non-convex combinatorial optimization problem. This inherently suffers from circular dependence and statistical inefficiency. Accurate clustering requires access to the underlying causal structures, while structure estimation depends on correct clustering—creating a self-reinforcing loop that is difficult to break without external supervision. Compounding this challenge is the statistical inefficiency in jointly navigating two exponentially large solution spaces—causal graphs and cluster assignments. In high-dimensional and limited-sample regimes, this leads to severe estimation variance, making reliable inference virtually impossible without additional constraints. Moreover, the mixed variable scenario (discrete and continuous variables coexist) creates conflicts, where causal structure learning tends to favor continuous variables due to their smoothness and differentiability, while clustering may be dominated by discrete variables that disproportionately influence similarity metrics. These opposing forces distort the optimization landscape, making it difficult for converging to a valid solution. Breaking this barrier typically requires auxiliary information—such as temporal ordering [33], environmental labels [21],[34],[35], or interventions [22]—to either anchor the clustering process or constrain the causal structures. Without such guidance, existing approaches struggle to reliably achieve this joint estimation task.

Second, the persistent inability to reconcile invariance and heterogeneity reflects a structural bias–variance antinomy, rooted in a conflict between the identifiability assumptions required for causality and the choice of modelling granularity necessary to capture meaningful heterogeneity. Achieving stable and transferable causal structures across contexts demands strong invariance assumptions, such as shared parameters [23] or mechanism constraints [24], which promote robustness but smooth over context-specific causal variations, potentially leading to underfitting. Conversely, relaxing these assumptions to capture finer-grained diversity inflates the model's effective degrees of freedom, but pushes estimation into high-variance regimes, making the model susceptible to overfitting, as bias can easily be mistaken for true causal shifts. Crucially, observational data alone offer no guidance on the appropriate level of granularity where causal mechanisms remain sufficiently stable while preserving genuine heterogeneity. This barrier is further amplified in mixed variable scenario. Mixed variables introduce incompatible modelling scales, forcing models to implicitly prioritize one variable type. Efforts to enhance invariance may suppress genuine heterogeneity driven by discrete variables, while capturing heterogeneity risks overfitting to spurious patterns amplified by variable-type imbalances. As a result, existing approaches are constrained to oscillate between robustness across contexts and fidelity within them.

Third, the presence of unmeasured confounders and data bias complicates the discovery of causal heterogeneity. They break the core assumptions underlying most identifiability results, rendering causal structures non-identifiable and blur the distinction between true heterogeneity and spurious variation. Unmeasured confounders can absorb dependencies among observed variables, reducing the constraints that data can place on the graph structure. Other biases further collapse informative statistical asymmetries, masking the signals need to infer causal directions or structure changes. Crucially, these sources of bias induce distributional shifts that are observationally equivalent to those generated by genuine causal heterogeneity. As a result, without auxiliary information [36-38] or stringent assumptions such as identically distributed variables [39], non-Gaussianity [40] or strong structure priors [41], it is difficult to determine whether observed heterogeneity reflects interpretable causal patterns or artifacts of latent bias. However, such information or assumptions are often unrealistic in real-world scenarios; for example, temporal information is typically unavailable, and mixed-type data do not satisfy distributional homogeneity.

### *B. Contribution*

To address the aforementioned challenges of heterogenous causal structure learning, we propose a novel unsupervised framework termed Interpretable Causal Mechanism-Aware Clustering with Adaptive Heterogeneous Causal Structure Learning (HCL). In contrast to prior models constrained by strong assumptions or reliance on external information, HCL is designed to discover heterogeneous causal patterns and their underlying mechanisms from scratch, directly from observational data without requiring temporal ordering, environment annotations, or interventions. HCL adaptively identifies both universal causal invariants and latent context-specific deviations, enabling end-to-end modeling of real-world causal heterogeneity while mitigating biases. Under mild assumptions, we theoretically prove that true structural heterogeneity is identifiable under the proposed framework. Across extensive experiments, HCL exhibits superiority in both causal structure learning accuracy and interpretable cluster recovery. Notably, validation on the single cell protein perturbation datasets show that the learned causal patterns align with known biological conditions and molecular interventions, supporting the biological plausibility and interpretability of the discovered structures.

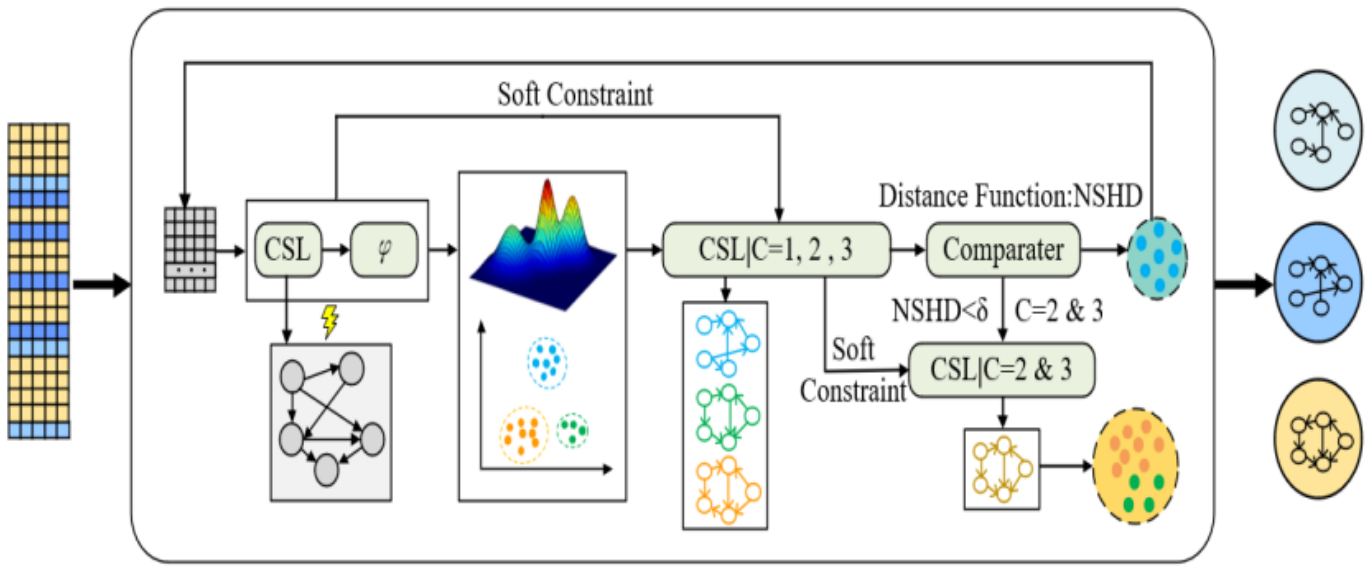

Fig. 1. Framework of HCL.

The main contributions of this work are as follows:

1) We propose the first unsupervised framework that jointly performs causal mechanism-aware clustering and heterogeneous causal structure learning from purely observational data. The framework relaxes the causal homogeneity assumption and supports mixed-type incomplete data, extending classical SCMs to better capture real-world data complexity.
2) At its core, we propose a structure-aligned equivalent representation, capturing causal semantics. This representation serves as an interpretable embedding space that enables causal semantic-level discrimination and facilitates the disentanglement of modifying and confounding effects.
3) Building on this foundation, we propose a bi-directional iterative strategy that jointly refines clustering, heterogenous causal structures, and confounding suppression. Without requiring prior knowledge of cluster number or graph topology, the strategy adaptively uncovers causal heterogeneity while attenuating latent biases. It ensures convergence to interpretable clusters aligned with genuine causal mechanisms and their modifying environments.
4) Additionally, we introduce a novel self-supervised regularization which enables flexibly balancing causal consistency and environment-specific modification, thus suppressing data bias and enhancing generalization.

## II. Methods

### A. Problem Formulation

Within the SCM framework, the causal mechanisms among a set of observed variables $X = (X_1, X_2, \ldots, X_D)$ are characterized by a DAG $G$ and a corresponding set of structural equations. Specifically, each variable $X_d$ is generated as:

$$X_d := f_d\left(\left\{X_h : h \in PA_d^G\right\}, U_d\right), d = 1, \ldots, D,$$

where $PA_d^G$ denotes the parent set of $X_d$ in $G$, $f_d$ is a deterministic function, and $U_d$ is an unobserved exogenous variable. Unlike an algebraic equation, the symbol ":=" specifies a directed causal assignment.

Two critical assumptions are typically imposed to enable identifiability and tractable learning. First, the variables $U = (U_1, \ldots, U_D)$ is assumed to be jointly independent, leading to the causal sufficiency condition:

$$P(U) = \prod_{d=1}^{D} P(U_d).$$

Moreover, under the causal homogeneity assumption, all samples $\{x^{(n)}\}_{n=1}^{N}$ are generated from the same fixed SCM, implying a shared, invariant causal graph:

$$G^{(1)} = G^{(2)} = \ldots = G^{(N)} = G.$$

Under these assumptions, the joint distribution $P(X)$ satisfies the causal Markov condition and is faithful to the underlying graph $G$. Thus, causal structure is theoretically identifiable from observational data.

However, real-world scenarios often violate these restrictive assumptions, especially when data are collected from heterogeneous contexts. Here, we explicitly relax the causal homogeneity assumption. Specifically, we assume that the observational dataset contains $N$ samples originating from $K \ll N$ distinct SCMs, each associated with a unique DAG $G^k$, such that

$$G^1 \neq G^2 \neq \ldots \neq G^K.$$

Each sample belongs to one of these $K$ latent groups, and shares the corresponding $G^k$. Each $G^k$ represents a distinct structural configuration, reflecting meaningful subgroup-specific causal structures.

In this heterogeneous setting, the differences among causal structures are modeled as modification effects arise from unobserved modifiers, characterized by the presence or absence of edges controlled by enhanced nodes [42]. Thus, modification effects represent genuine and interpretable structural variations. In contrast, confounding effects arise from unobserved common causes [3] and manifest as spurious statistical associations rather than true causal connections. While modification effects alter the DAG structure itself, confounding effects introduce misleading statistical dependencies without altering the underlying causal structure.

In this work, interpretable clustering is defined to recover the latent number $K$ of distinct causal subgroups with their corresponding subgroup-specific causal structures $G^k$ implying heterogeneity mechanism. Heterogeneous causal structure learning aims to preserve true structural modification effects and eliminate confounding or other noise-induced false discoveries of structural configuration. While each $G^k$ characterizes a distinct causal structure corresponding to a specific subgroup, different $G^k$ may still share a subset of common causal relationships. This reflects the presence of universal causal mechanisms that remain invariant across contexts.

Then, under the heterogeneous setting, we consider mixed-type observational data consisting of both discrete and continuous variables. We adopt the extended linear causal model (ELCM) [17] to explicitly specify the structural equations underlying the mixed data, such that:
Each continuous variable $X_j$ is assumed to be generated linearly from its parents specified by $G^k$:

$$X_j := \boldsymbol{\beta}_j{}^T X_{PA_j^{G^k}} + U_j, U_j \sim \mathrm{P}\left(U_j\right). \quad (1)$$

Each discrete variable $X_j$ is assumed to be generated by thresholding a linear combination of its parents specified by $G^k$:

$$X_j := \begin{cases} 1, \boldsymbol{\beta}_j{}^T X_{PA_j^{G^k}} + U_j > 0 \\ 0, \boldsymbol{\beta}_j{}^T X_{PA_j^{G^k}} + U_j \le 0 \end{cases}, U_j \sim \mathrm{P}\left(U_j\right). \quad (2)$$

Here, $X_{PAj}{}^{Gk} \in \mathbb{R}^p$ collects its parent variables, $\boldsymbol{\beta}_j{}^T \in \mathbb{R}^{1\times p}$ is the corresponding row coefficient vector. The structural coefficients can be organized into a weighted adjacency matrix $\mathbf{B} = \left[b_{ij}\right] \in \mathbb{R}^{d\times d}$, where $b_{ij}$ quantifies the direct causal effect of $X_i$ on $X_j$. For each *j*, the vector $\boldsymbol{\beta}_j$ is precisely the subvector of the *j*-th column of $\mathbf{B}$ restricted to the parent set, i.e.,$b_{ij} \neq 0$ indicates the presence of a directed edge $X_i \to X_j$, whereas $b_{ij} = 0$ indicates its absence.

We aim to simultaneously identify interpretable clusters and heterogenous causal structures underlying such mixed data without prior knowledge such as pre-defined cluster numbers and fixed skeletons.

### *B. Overview*

We propose HCL to address the aforementioned challenges. It functions under non-SCM-conforming data, where the two critical assumptions are violated due to heterogeneity or confounding. HCL is designed to reveal latent heterogeneity of causal structures caused by unobserved modifiers and mitigate spurious associations caused by unobserved confounders in an unsupervised manner. It consists of two principal components:

**Causal Structure Heterogeneity Representation** (Section II.C):

To characterize both unobserved modifiers and confounders, we construct an equivalent model that jointly encodes implicit causal structure heterogeneity and potential confounding effects. Specifically, we define the structure-aligned latent variables via a mapping function grounded in the shared causal backbone, and derive their analytical formulation, which enables the projection of heterogeneous and spurious causal edges into a latent representation space. It forms the basis for both clustering and structure learning, supporting inference even in mixed-type data by capturing multi-distribution variables within a unified mapping formulation.

Theoretically, we prove that the latent causal structure heterogeneity and spurious effects can be identified from the equivalent representation. A counterintuitive but theoretically guaranteed result is derived: injecting a common structure prior can enhance the identifiability of causal structural heterogeneity.

**Iterative Learning Strategy** (Section II.D):

To uncover latent clusters of causal structural heterogeneity in a data-adaptive manner, we design an alternating optimization procedure that iteratively executes three sub-steps: causal structure learning, equivalent latent variable extraction, and sample-wise clustering. This cycle continues until convergence. Through this mechanism, HCL achieves two critical goals. First, it simultaneously discovers clusters with distinct yet explainable causal structures, enhancing interpretability of heterogeneous groups and preserving modifying effects while suppressing confounding biases. Second, it jointly estimates both the universal and cluster-specific causal structures, without prior specification of the number of clusters, external information or stringent assumptions.

### *C. Causal Structure Heterogeneity Representation*

To formally capture the heterogeneity of causal structures, we propose a latent representation that projects observed data into a space where distinct causal structures are more separable.

1) Model Assumptions and Notation

Let $X = (X_1, X_2, \ldots, X_D)$ denote a set of $D$ observed variables, where each $X_j$ may be continuous or discrete. The data is assumed to be generated from multiple underlying causal mechanisms. We denote the latent cluster index as $C \in \{1, \ldots, K\}$, which represents the underlying causal structure to which a sample belongs. However, $C$ is not observed and $K$ is not pre-defined.

Each cluster follows a distinctive ELCM:

$$X_j = \mathrm{f}_j^c\left(PA_j^c, U_j^c\right), U_j^c \sim \mathrm{P}\left(\theta^c\right),$$

where $PA_j^c \subset X \backslash \{X_j\}$ are the parents of $X_j$ under cluster $c$, and $P(\theta^c)$ is a Gaussian distribution independent of $PA_j^c$. The observed data is a mixture of these mechanisms:

$$X \sim \sum_{c=1}^{K} \pi_c \, \mathrm{P}^c\left(X\right),$$

where $\pi_c = \mathrm{P}(C = c)$ and $C \sim \mathrm{Categorical}(\pi_1, \ldots, \pi_K)$.

2) Definition of Equivalent Latent Variable

We define the equivalent latent variables $Z = (Z_1, Z_2, \ldots, Z_D)$ as a continuous proxy of the latent index $C$, learned from the sample-wise causal reconstruction discrepancy under the shared causal backbone $G_c$.

Let $\widehat{X}_j = \widehat{\mathrm{f}}_j\left(\widehat{PA}_j\right)$ denote the estimated value of $X_j$ under $G_c$. Define $Z_j = \mathrm{E}\left[\widehat{U}_j \middle| X_j, \widehat{X}_j, \theta\right] = \varphi\left(X_j, \widehat{X}_j, \theta\right)$, where $\varphi(\cdot)$ is a parameterized function that maps the mixed data to a continuous latent representation. $\widehat{U}$ is the independent and identically distributed proxy of the exogenous variables including latent modifiers $M$ and confounders $L$.

Under our model assumptions, we derive the analytical formulation of $\varphi(\cdot)$:
If $X_j$ is continuous variable:

$$\varphi\left(X_j, X_j, \boldsymbol{\theta}\right) = \{\hat{\boldsymbol{\mu}} + \boldsymbol{\Sigma}(\mathbf{I} - \hat{\mathbf{B}}^{\mathrm{T}})^{-\mathrm{T}}\left[(\mathbf{I} - \hat{\mathbf{B}}^{\mathrm{T}})^{-1}\boldsymbol{\Sigma}(\mathbf{I} - \hat{\mathbf{B}}^{\mathrm{T}})^{-\mathrm{T}}\right]^{-1} (X^T - \left(\mathbf{I} - \hat{\mathbf{B}}^{\mathrm{T}})^{-1}\hat{\boldsymbol{\mu}}\right)\}_j = X_j - X_j;$$

If $X_j$ is discrete variable:

$$\varphi\left(X_j, X_j, \theta\right) = \begin{cases} \hat{\mu}_j - \dfrac{\hat{\sigma}_j}{\sqrt{2\pi}(1-\rho)} \exp\left(\dfrac{-\left(X_j + \hat{\mu}_j\right)^2}{2\hat{\sigma}_j{}^2}\right), \text{if } X_j = 0; \\ \hat{\mu}_j + \dfrac{\hat{\sigma}_j}{\sqrt{2\pi}\rho} \exp\left(\dfrac{-\left(X_j + \hat{\mu}_j\right)^2}{2\hat{\sigma}_j^2}\right), \text{if } X_j = 1. \end{cases} \quad (3)$$

Here, $\boldsymbol{\theta} = \left(\widehat{\boldsymbol{\mu}}, \widehat{\boldsymbol{\Sigma}}\right)$, where $\widehat{\boldsymbol{\mu}} \in \mathbb{R}^D$ and $\widehat{\boldsymbol{\Sigma}} \in \mathbb{R}^{D\times D}$ denote the mean vector and covariance matrix of the Gaussian distribution, respectively; and $\rho = \Phi\left(\frac{\widehat{X_j} + \hat{\mu}_j}{\hat{\sigma}_j}\right)$.

3) Identifiability of Structural Heterogeneity

We show that, under mild conditions, the proposed framework suffices to identify structural heterogeneity.

**Proposition1:** Suppose the observational data $\chi$ is generated from a mixture of heterogeneous structural causal model indexed by $M$, and further influenced by $L$. Denote the whole causal structure of both observed and latent variables as $O \in \Omega$. There exists a non-trivial SCM $S$ faithful to $G^c$, estimated from $\chi$. Then: (1) there exist $O_i \neq O_j$ that yield the same $p(X)$; (2) there not exist $O_i \neq O_j$ that yield the same $p(Z|X,S)$.

We prove **Proposition 1** in Supplementary Material Section A. According to this, the causal structure cannot be consistently recovered from $p(X)$ alone. In contrast, the causal structure is identifiable based on such posterior expectation $Z$ inferred one-to-one from $p(L,M|X,S)$. Thus, $Z$ serves as an informative embedding to separate different causal structure or confounding modes. Intuitively, according to $Z$, samples can be clustered into distinct modes implying different $M$-indexed causal structures and within each cluster, the effect of confounders is controlled thus removes much of the spurious association.

**Proposition2:** Adding a penalty that favors a shared causal backbone $G_c$ (i) suppresses spurious heterogeneity and (ii) accentuates genuine heterogeneity.

**Proof**: Assume in cluster $k$,

$$\hat{\mathbf{B}}^{k,c} = \underset{\mathbf{B}}{\operatorname{argmin}}[L_k(\mathbf{B}) + \lambda_1 \sum_{ij \in G^c} |\mathbf{B}_{ij}| + \lambda_2 \sum_{ij \notin G^c} |\mathbf{B}_{ij}|], \lambda_2 > \lambda_1 \geq 0 .$$

$$\hat{\mathbf{B}}^{k,o} = \underset{\mathbf{B}}{\operatorname{argmin}}[L_k(\mathbf{B}) + \lambda \sum_{ij} |\mathbf{B}_{ij}|], \lambda_1 < \lambda < \lambda_2.$$

Here, $\lambda_1 < \lambda < \lambda_2$ is a necessary condition for these two methods to estimate a consistent number of positive edges. For any edge $ij \notin G^k$. By the central limit theorem,

$$\frac{\partial L_k}{\partial \mathbf{B}_{ij}} \sim \mathrm{N}\left(0, \frac{\sigma^2}{n_k}\right),$$

$$\mathrm{P}\left(\hat{\mathbf{B}}_{ij}^{k,c} \neq 0 \mid \mathbf{B}_{ij}^k = 0\right) = 2\Phi\left(-\frac{\sqrt{n_k}\lambda_*}{\sigma}\right), \lambda_* = \begin{cases} \lambda_1, \text{if } ij \in G^c \\ \lambda_2, \text{if } ij \notin G^c \end{cases},$$

$$\mathrm{P}\left(\hat{\mathbf{B}}_{ij}^{k,o} \neq 0 \mid \mathbf{B}_{ij}^k = 0\right) = 2\Phi\left(-\frac{\sqrt{n_k}\lambda}{\sigma}\right).$$

Similarly, let $ij$ be a true cluster-specific edge that exists only in cluster $k$, then

$$\frac{\partial L_k}{\partial \mathbf{B}_{ij}} \sim \mathrm{N}\left(\mathbf{B}_{ij}^k, \frac{\sigma^2}{n_k}\right),$$

$$\mathrm{P}\left(\hat{\mathbf{B}}_{ij}^{k,c} \neq 0 \mid \mathbf{B}_{ij}^k \neq 0\right) = 1 - \Phi\left(\frac{\lambda_* - \mathbf{B}_{ij}^k}{\sigma / \sqrt{n_k}}\right) + \Phi\left(\frac{-\lambda_* - \mathbf{B}_{ij}^k}{\sigma / \sqrt{n_k}}\right),$$

$$\mathrm{P}\left(\hat{\mathbf{B}}_{ij}^{k,o} \neq 0 \mid \mathbf{B}_{ij}^k \neq 0\right) = 1 - \Phi\left(\frac{\lambda - \mathbf{B}_{ij}^k}{\sigma / \sqrt{n_k}}\right) + \Phi\left(\frac{-\lambda - \mathbf{B}_{ij}^k}{\sigma / \sqrt{n_k}}\right);.$$

For any $s \neq k$:

$$\mathrm{P}\left(\hat{\mathbf{B}}_{ij}^{s,c} = 0 \mid \mathbf{B}_{ij}^s = 0\right) = 1 - 2\Phi\left(-\frac{\sqrt{n_s}\lambda_*}{\sigma}\right), \lambda_* = \begin{cases} \lambda_1, \text{if } \ ij \in G^c \\ \lambda_2, \text{if } \ ij \notin G^c \end{cases},$$

$$\mathrm{P}\left(\hat{\mathbf{B}}_{ij}^{s,o} = 0 \mid \mathbf{B}_{ij}^s = 0\right) = 1 - 2\Phi\left(-\frac{\sqrt{n_s}\lambda}{\sigma}\right).$$

Here, $n_k, n_s$ denote the sample sizes of cluster $k, s$, respectively. Let $\widehat{H}_C^k$ be the estimated edges set of cluster $k$ using $G^c$, $\widehat{H}_O^k$ the estimated edges set of cluster $k$ without $G^c$, and $H^k$ the true edges set of cluster $k$. Define the estimated false edges set as $\hat{V}_*^k := \widehat{H}_*^k \backslash H^k$, and the false discovery rate as

$$FDR_*^k := \frac{E\left[\left|\hat{V}_*^k\right|\right]}{\left|\hat{H}_*^k\right|} \propto \frac{E[\hat{B}_{ij}^{k,*} \neq 0 \mid B_{ij}^k = 0]}{\left|\hat{H}_*^k\right|}.$$

Assume that $\gamma = \frac{|G^c \backslash H^k|}{D^2}$ and $|\widehat{H}_c^k| = |\widehat{H}_o^k|$. Denote $\Phi\left(-\frac{\sqrt{n_k}\Lambda}{\sigma}\right)$ as $\rho(\Lambda)$. Then we have:

$$\frac{\mathrm{E}\left[\left|\hat{V}_c^k\right|\right]}{\mathrm{E}_\lambda\left[\mathrm{E}\left[\left|\hat{V}_o^k\right|\right]\right]} = \frac{\gamma\rho(\lambda_1) + (1-\gamma)\rho(\lambda_2)}{\rho(\lambda^\varepsilon)},$$

where $\rho(\lambda^\varepsilon) = \frac{1}{\lambda_2 - \lambda_1}\int_{\lambda_1}^{\lambda_2} \rho(\lambda) \mathrm{d}\lambda$, $\lambda_1 < \lambda^\varepsilon < \lambda_2$, according to mean value theorem of integrals. Since $\rho(\Lambda)$ is a monotonically decreasing convex function of $\Lambda$ and according to Hadamard inequality:

$$\rho(\lambda^\varepsilon) \geq \rho\left(\frac{\lambda_1 + \lambda_2}{2}\right).$$

According to Lagrange's mean value theorem,

$$\begin{aligned} &\gamma\rho(\lambda_1) + (1-\gamma)\rho(\lambda_2) = \rho(\lambda_2) + \gamma(\rho(\lambda_1) - \rho(\lambda_2)) \\ &= \rho(\lambda_2) + \gamma\rho'(\xi)(\lambda_1 - \lambda_2), \lambda_1 < \xi < \lambda_2. \end{aligned}$$

According to Taylor's expansion:

$$\rho\left(\frac{\lambda_1 + \lambda_2}{2}\right) = \rho(\lambda_2) + \rho'(\lambda_2)\frac{\lambda_1 - \lambda_2}{2} + \frac{\rho''(\epsilon)}{2}(\lambda_1 - \lambda_2)^2,$$

where $\frac{\lambda_1 + \lambda_2}{2} < \epsilon < \lambda_2$. For sparsity, $0 < \gamma \ll \frac{1}{2}$. Thus,

$$\gamma|\rho'(\xi)| < \frac{|\rho'(\lambda_2)|}{2}.$$

Additionally, $\rho''(\epsilon) > 0$. Thus,

$$\gamma\rho(\lambda_1) + (1-\gamma)\rho(\lambda_2) < \rho\left(\frac{\lambda_1 + \lambda_2}{2}\right) \leq \rho(\lambda^\varepsilon).$$

Thus,

$$E\left[\left|\hat{V}_c^k\right|\right] < E\left[\left|\hat{V}_o^k\right|\right].$$

Thus, (i) is proved.

Denote the heterogenous edges between clusters $k$ and $s$ as $Het(H) := H^k \backslash H^s$, and the estimated true heterogenous edges set $Het(\widehat{H}) \cap Het(H)$ as $\hat{T}_*$. For true cluster -specific edge $B_{ij}^k$ in cluster $k$ and not in other clusters, denote the probability $P_c(\hat{B}_{ij}^{k,*} \neq 0 | B_{ij}^k \neq 0) P_c(\hat{B}_{ij}^{s,*} = 0 | B_{ij}^s = 0)$ of it be estimated specifically as $s(\Lambda)$:

$$s(\Lambda) = \left[1 - \Phi\left(\frac{\Lambda - B_{ij}^k}{\sigma / \sqrt{n_k}}\right) + \Phi\left(\frac{-\Lambda - B_{ij}^k}{\sigma / \sqrt{n_k}}\right)\right]\left[1 - 2\Phi\left(-\frac{\sqrt{n_s}\Lambda}{\sigma}\right)\right].$$

Similarly, we can prove:

$$\frac{E\left[\left|\hat{T}_c\right|\right]}{E_\lambda\left[E\left[\left|\hat{T}_o\right|\right]\right]} = \frac{\gamma s(\lambda_1) + (1-\gamma)s(\lambda_2)}{s(\lambda^s)} > 1,$$

where $s(\lambda^s) = \frac{1}{\lambda_2 - \lambda_1}\int_{\lambda_1}^{\lambda_2} s(\lambda) \mathrm{d}\lambda$, $\lambda_1 < \lambda^s < \lambda_2$. Thus, (ii) is proved.

**Proposition 3:** Clustering is performed on $Z$. $\widehat{C^n}$ and $\widehat{G^C}$ are obtained via the bi-directional strategy within the HCL framework. Assume the true causal graphs are distinct:

$$\delta_{\min}^{\mathrm{SHD}} = \min_{k \neq l} \mathrm{SHD}(G^k, G^l) > 0. \quad \text{(A1)}$$

If the estimated graph $\widehat{G^C}$ of a cluster $c$ containing samples primarily from one generating mechanism $G^k$ satisfies:

$$d_{\mathrm{SHD}}\left(G^{C},G^{k}\right)\xrightarrow{\mathrm{P}}0 \text{ as } N_{c}\to\infty\,;$$

then the final cluster label $\widehat{C^n}$ converges to the true label $C^n$:

$$\lim_{N\to\infty} \mathrm{P}(C^{n}=k|C^{n}=k)=1\,.$$

For the proof of **Proposition 3**, see the Supplementary Material Section B.

*D. Iterative Learning Strategy*

**(1) Self-supervised causal backbone regularization**

The global patterns learned from earlier rounds serve as soft constraint for the current iteration, propagating structural consistency across iterations. Specifically, we incorporate a shared causal backbone derived from the parent cluster of each subcluster to enhance the estimation of subcluster-specific causal structures. This mechanism helps preserve transferable causal edges while allowing meaningful deviations. Formally, let $G_t^k$ represent the subcluster-specific causal structure of the $k$-th subcluster at iteration $t$, and $G_{t-1}^c$ denote the shared causal backbone estimated from the previous iteration. The optimization objective function at iteration $t$ is then defined as the sum of a data-fitting term and a soft constraint term, i.e.,

$$L\left(G_t^k,D^k \mid G_{t-1}^c\right)=L\left(G_t^k,D^k\right)+\lambda R_{\mathrm{CCE}}\left(G_t^k \mid G_{t-1}^c\right).$$

Here, following [17], $L(G_t^k, D^k)$ denotes the negative log-likelihood of the causal structure given the subcluster-specific data $D^k$, while $R_{\mathrm{CCE}}(G_t^k \mid G_{t-1}^c)$ represents the soft constraint, balancing universal causal patterns and specific variations, defined as:

$$\lambda_1 \sum_{ij\in G_{t-1}^c}\left|(\mathbf{B}_t^k)_{ij}\right|+\lambda_2 \sum_{ij\notin G_{t-1}^c}\left|(\mathbf{B}_t^k)_{ij}\right|,\lambda_1<\lambda_2. \tag{4}$$

Here, the penalty coefficients assigned to edge $(i,j)$ is smaller if it appears in the parent-derived backbone, whereas a larger penalty is imposed if the edge is absent from the backbone.

**(2) Dirichlet prior Bayesian Gaussian mixture model**

We assume Dirichlet prior over cluster weights as $p$~Dirichlet($\alpha$), and assume:

$$Z^n \in \mathbb{R}^D,$$

$$Z^n \mid C^Z = i \sim \mathrm{N}\left(\mu_i^Z,\Sigma_i^Z\right),$$

$$p\left(Z^n\right)=\sum_{i=1}^{I} p_i\,\mathrm{N}\left(\mu_i^Z,\Sigma_i^Z\right),$$

$$\mu_i^Z \sim \mathrm{N}\left(m_0,\beta_0^{-1}I\right),\Sigma_i^{Z-1}\sim \mathrm{W}\left(v_0,W_0\right).$$

Here, $\mu_i^Z \in \mathbb{R}^D$ and $\sum_i^Z \in \mathbb{R}^{D\times D}$ denote the mean and covariance of the *i*-th Gaussian component, and $\alpha$ is the hyperparameter vector of the Dirichlet prior controlling the sparsity of cluster proportions. This expression defines a soft cluster assignment for each sample, reflecting uncertainty in the inferred causal mechanism identity. The incorporation of the Dirichlet prior allows for adaptive control of cluster concentration and supports automatic relevance determination in the presence of redundant or spurious clusters. The posterior is approximated via variational inference:

$$q(p,C^Z,\{\mu_i^Z,\Sigma_i^Z\})=q(p)q(C^Z)\prod_i q(\mu_i^Z,\Sigma_i^Z)\,.$$

We maximize the ELBO:

$$L\left(q\right)=\mathrm{E}_q[\log p(\mathbf{Z},C^Z,\pi,\mu,\Sigma)]-\mathrm{E}_q[\log q(C^Z,\pi,\mu,\Sigma)]\,.$$

During inference, we alternate between computing soft cluster responsibilities:

$$r_{ni}=q(C^Z=i)\propto \exp\{\mathrm{E}_q[\log p_i]+\mathrm{E}_q[\log \mathrm{N}(\mu_i^Z,\Sigma_i^Z)]\}\,,$$

where:

$$\mathrm{E}_q\left[\log p_i\right]=\varphi\left(\alpha_i\right)-\varphi(\sum_{i'}\alpha_{i'})\mathrm{E}_q[\log \mathrm{N}(\mu_i^Z,\Sigma_i^Z)]=$$

$$-\left(D/2\right)\log\left(2\pi\right)+\left(1/2\right)\left[\sum_{d=1}^{D}\psi\left(\left(v_i+1-d\right)/2\right)+D\log\left(2\right)+\log\left|W_i\right|\right]$$

$$-\left(1/2\right)\left[D/\beta_i+v_i\left(Z^n-m_i\right)^T W_i\left(Z^n-m_i\right)\right]$$

and updating the posterior distribution parameters based on the responsibilities:

$$\alpha_i=\alpha+N_i\;,N_i=\sum_n r_{ni}\,,\beta_i=\beta_0+N_i\;,m_i=\frac{\beta_0 m_0+N_i\overline{Z^i}}{\beta_i}\,,v_i=v_0+N_i\,,$$

$$W_i^{-1}=W_0^{-1}+S_i+\frac{\beta_0 N_i}{\beta_0+N_i}(\overline{Z^i}-m_0)(\overline{Z^i}-m_0)^T\,,\overline{Z^i}=\frac{1}{N_i}\sum_n r_{ni}Z^n\,,$$

$$S_i=\sum_n r_{ni}Z^n-\overline{Z^i}(Z^n-\overline{Z^i})^T\,.$$

This inference procedure is iterated until convergence. The resulting cluster labels:

$$C_n^Z=\mathrm{argmax}_i\; r_{ni} \tag{5}$$

are then used to guide cluster-specific causal structure learning in the next iteration.

**(3) Bi-directional strategy**

In contrast to traditional hierarchical clustering methods that follow either a purely top-down or bottom-up procedure, we propose a bi-directional strategy that obviates the need for pre-specifying the number of clusters. Specifically, our approach iteratively alternates between mutually reinforcing top-down and bottom-up steps to simultaneously refine heterogenous causal structures and optimize clustering results: (1) clustering samples into interpretable subgroups based on their latent representations and learning subgroup-specific causal structures conditioned on current cluster assignments; (2) refining cluster assignments and subgroup-specific causal structures based on structural similarity. This strategy allows for bootstrapping from a structure-free initialization and progressively converging toward a stable partitioning that aligns with genuine modifying environments and their corresponding causal graphs. Crucially, it also allows confounding suppression without explicitly modeling latent confounders—by iteratively isolating and reducing spurious dependencies in the latent space.

In the top-down phase, we compute the posterior expectation of latent variables for each sample based on the shared causal backbone derived from the previous iteration, and subsequently perform Bayesian Gaussian mixture model clustering with a Dirichlet prior to obtain preliminary subclusters. Within each subcluster, subcluster-specific causal structures are inferred using procedure detailed in **Section D (1)**. Following this subcluster partitioning and inference, we assess pairwise structural similarities between subclusters via Structural Hamming Distance (SHD) normalized by the average number of inferred edges, and merge subclusters based on their structural similarities. Then, the merged structure is recomputed according to weighted average matrix **W** with each element $w_{ij}$ defined as:

$$w_{ij}=\sum_{m=1}^{M}\frac{n_m}{n_T}\cdot 1\left\{\left(i,j\right)\in G_m\right\}$$

where $M$ is the number of subclusters need to be merged, $n_m$ is the sample size belonging to $m$-th subcluster, $n_T$ is the total sample size of the merged cluster. $1\{(i,j) \in G_m\}$ is the indicator function equaling to 1 if edge $(i,j)$ exists in $m$-th structure and 0 otherwise. Then, we refit the merged adjacency matrix $\mathbf{B}_M$ use the continuous variant of **Section D (1)** by:

$$\sum_{ij} \lambda_{ij} \left|(\mathbf{B}_M)_{ij}\right|, \quad \lambda_{ij} = \frac{1}{1+\exp\left(\eta\left(w_{ij}-\tau\right)\right)} \tag{6}$$

where $\eta$>0 controls the sharpness of the transition and $\tau \in (0,1)$ denotes the threshold at which the penalty is approximately 0.5. In our implementation, we set $\eta = 20$ and $\tau$=0.5 to enforce strong penalization for weakly supported edges and negligible penalization for edges with strong cross-cluster consensus.

The refinement process proceeds in a bottom-up manner, propagating upward to revise the parent-level structures and reconfigure the hierarchical cluster assignments across different levels within the current iteration. In the current iteration, unmerged sample clusters may exhibit greater structural heterogeneity and are thus carried forward into the next iteration. If parent-level clusters are merged, their corresponding subclusters are discarded. For each newly merged or reassigned cluster, a new cluster-specific causal structure is inferred, and each resulting structure is subsequently compared against all others to assess structural similarity and guide further reorganization.

The bi-directional procedure is executed iteratively until convergence, defined as the iteration at which no further novel subcluster-specific causal structures emerge. Upon convergence, the algorithm outputs the final cluster assignments and their corresponding subcluster-specific causal structures.

**Algorithm 1:** Heterogenous Causal Structure Learning

**Input:** Mixed observational data $\chi$; structural similarity threshold $\delta$; penalty coefficient $(\lambda_1, \lambda_2)$; maximum number of iteration $T$.

**Output:** Cluster number $K$, cluster assignment $\mathbf{C}_K$, cluster-specific causal graph adjacency matrices $\mathbb{B}_t = \{\mathbf{B}_{t1}, \mathbf{B}_{t2}, \dots, \mathbf{B}_{tK}\}$.

**Initialize:** $t = 1$; $\mathbf{C}_K = \mathbf{0} \in \mathbb{R}^n$ ; $\mathbf{C}_t = \mathbf{0} \in \mathbb{R}^n$ ; compute initial backbone [17] and use it as initial $\mathbb{B}_t$ .

**Repeat**
- **for** each cluster in $\mathbf{C}_t$ **do**
  - Calculate $\mathbf{Z}$ using Eq. (3);
  - calculate subcluster assignment $\mathbf{S}_c$ using Eq. (5);
  - **for** each subcluster in $\mathbf{S}_c$ **do**
    - Infer subcluster-specific adjacency matrix $\mathbf{B}_S$ using Eq. (4);
  - **end for**
  - Compute NSHD between $\mathbf{B}_S$; merge if NSHD $\leq \delta$;
  - recompute $\mathbf{B}_M$ using Eq. (6);
- **end for**
- Adjust $\mathbb{B}_t$ by $\mathbb{B}_S$ using Eq. (6);
- recompute $\mathbb{B}_t$ after computing NSHD and merging;
- dissolve associated subclusters of merged clusters;
- $t \leftarrow t + 1$;
- update $\mathbf{C}_t$ and $\mathbb{B}_t$ if identifying new subcluster;
- reassign and recompute $\mathbf{C}_t$ , $\mathbb{B}_t$ after computing NSHD and merging;
- $\mathbf{C}_K[i] \leftarrow \mathbf{C}_t[i], \forall i \in supp(\mathbf{C}_t)$;
- $\mathbf{C}_t \leftarrow$ non-merged clusters with potential heterogeneity

**until** convergence or $t > T$

$K \leftarrow | \{c \in \mathbf{C}_K\} |$

## III. EXPERIMENTS

To comprehensively evaluate the performance of HCL, we conduct experiments on both synthetic datasets with ground-truth causal graphs and real-world datasets. The primary goals of our evaluation are to (i) assess the capability of HCL to identify heterogeneous causal structures and cluster samples accordingly; (ii) examine its accuracy and robustness under a wide range of challenging conditions, including limited sample sizes, varying degrees of class imbalance, increasing graph complexity, and different numbers of underlying causal clusters; and (iii) demonstrate its interpretability in discovering distinct cluster-specific causal patterns.

### *A. Synthetic Datasets*

We generate synthetic datasets from a mixture of structural causal models with varying causal mechanisms. Each mechanism corresponds to a distinctive cluster. To be specific, we generate random DAGs as ground-truth causal graphs using ER model [43]. Edge weights are sampled independently and uniformly from the intervals [-2, -0.5] and [0.5, 2], ensuring sufficient edge strength while avoiding near-zero effects. This procedure produces weighted DAGs and their corresponding adjacency matrices **B**. Given each weighted DAG, we generate samples according to (1)-(2), where the exogenous variables are drawn from a standard Gaussian distribution. Each DAG and its corresponding data samples are labeled as belonging to a distinct causal class, providing the ground truth for evaluating both interpretable clustering and causal structure learning performance.

1) **Datasets 1**: Dataset 1 is designed to evaluate the influence of sample size on the performance. Each dataset instance contains two causal classes as ground-truth clusters, each corresponding to a distinct random weighted DAG with 10 nodes and 10 directed edges. Samples from each class are generated using the corresponding DAG. Across different dataset instances, the per-class sample sizes vary among 500, 300, and 200, allowing analysis of how decreased data availability affects both structure accuracy and clustering consistency.

2) **Datasets 2**: Dataset 2 is designed to assess robustness under class imbalance. Each dataset instance contains two causal classes, each represented by a distinct random weighted DAG with 10 nodes and 10 directed edges. The maximum number of samples per instance is fixed at 500, while the sample proportions between the two classes are varied across the following configurations: balanced setting ([500, 500]), mild imbalance ([100, 500]), and severe imbalance ([50, 500]). This design allows for simulating increasing imbalance severity while keeping graph complexity constant.

3) **Datasets 3**: Dataset 3 is designed to evaluate how the magnitude of causal graph structural discrepancy influences the performance. Each dataset instance contains two causal classes, where samples are generated from distinct random weighted 10-node DAGs. We vary the edge ratio of the two DAGs across three configurations: 5 edges vs. 10 edges, 10 edges vs. 10 edges, and 20 edges vs. 10 edges. In all configurations, each class contains 500 samples.

4) **Datasets 4**: Dataset 4 examines the impact of causal graph complexity on performance, by varying both the number of nodes and the density of edges. Each dataset instance contains two causal classes, each generated from a distinct random weighted DAG, with 500 samples per class. Four levels of graph complexity are considered: sparse (10 nodes, 5 edges), low (10 nodes, 10 edges), moderate (10 nodes, 20 edges), and high (20 nodes, 100 edges).

5) **Datasets 5**: Dataset 5 is designed to evaluate the scalability in handling an increasing number of underlying causal clusters. Additionally, it is also used to examine whether adaptive methods introduce spurious heterogeneity in cases where the underlying data are structurally homogeneous. Beyond this, Dataset 5 also serves to evaluate HCL's performance in causal structure learning under homogeneous settings, where HCL aims to mitigate spurious edges induced by confounding and noise. Each dataset instance contains $K \in \{1,3,5,7\}$ causal classes, where each class corresponds to a distinct random weighted DAG with 10 nodes and 10 edges. Each class contains 500 samples generated from its corresponding causal mechanism. The task becomes increasingly challenging as more causal graphs must be accurately disentangled from observational data. This evaluation also provides an important test of structural consistency, ensuring that the method does not over-segment data or detects artificial differences when genuine causal heterogeneity is absent.

## *B. Evaluation Metrics and Baselines*

We employ three widely-used metrics to evaluate the performance of HCL in both clustering and causal structure learning: Adjusted Rand Index (ARI), False Discovery Rate (FDR), and True Positive Rate (TPR). ARI is used to assess clustering quality by quantifying the agreement between inferred cluster assignments and ground-truth labels, adjusted for chance. It ranges from −1 to 1, with 1 indicating perfect clustering, 0 corresponding to random labeling, and negative values denoting worse-than-random assignments. A higher ARI indicates better clustering performance. FDR and TPR are used to evaluate the quality of causal structure learning; both metrics range from 0 to 1. Specifically, FDR measures the proportion of falsely identified edges among all predicted edges, with lower values indicating fewer false positives and hence better precision; TPR measures the proportion of true causal edges correctly recovered, with higher values indicating better recall. All metrics are computed per cluster and averaged over ten independent repetitions to ensure statistical stability.

For clustering evaluation, we compare HCL against representative baseline clustering methods. Specifically, we include Gaussian Mixture Model (GMM), KMeans [44], Variational Bayesian Gaussian Mixture Model (VBGMM) [45] which adopts a Dirichlet distribution prior, all of which require the number of clusters to be predefined. We additionally consider the Dirichlet Process clustering (DP) [46], which allows the number of clusters to be inferred from the data. We provide GMM, KMeans, and VBGMM with the exact number of clusters to ensure a fair comparison, and evaluate whether HCL achieves better clustering by exploiting structural heterogeneity beyond statistical similarities. When the number of classes exceeds two, we continue to provide the exact number to methods that require it, while testing the ability of DP and HCL to infer the number of clusters automatically. Notably, HCL performs binary splitting at each iteration and adaptively determines the final number of clusters based on the structure-induced convergence of causal patterns.

For causal structure learning, we compare HCL with NOTEARS and KEEL. NOTEARS is a widely used continuous optimization-based method for structure learning from observational data, applied here within each inferred cluster. KEEL is a weakly supervised causal discovery framework guided by fuzzy knowledge, designed to improve causal structure recovery in complex data. We adopt a purely observational data-driven setting and do not introduce any manually specified causal relations or structural priors. In our evaluation, we evaluate the performance of methods within each cluster by comparing the inferred structure against the corresponding ground-truth causal graph of that cluster. These comparisons allow us to assess HCL's distinct advantage in disentangling and recovering heterogenous causal structures. In addition, they provide insight into the extent to which existing structure learning methods can generalize across distinct causal mechanisms and to heterogeneous settings with multiple causal mechanisms.

## *C. Results for Synthetic Datasets*

All experiments are conducted under a consistent hyperparameter setting for HCL. Specifically, the hyperparameters are set as $\delta = 1$, $\lambda_1 = 0.1$, $\lambda_2 = 0$, and $T = 100$.

1) **Results of Datasets 1**

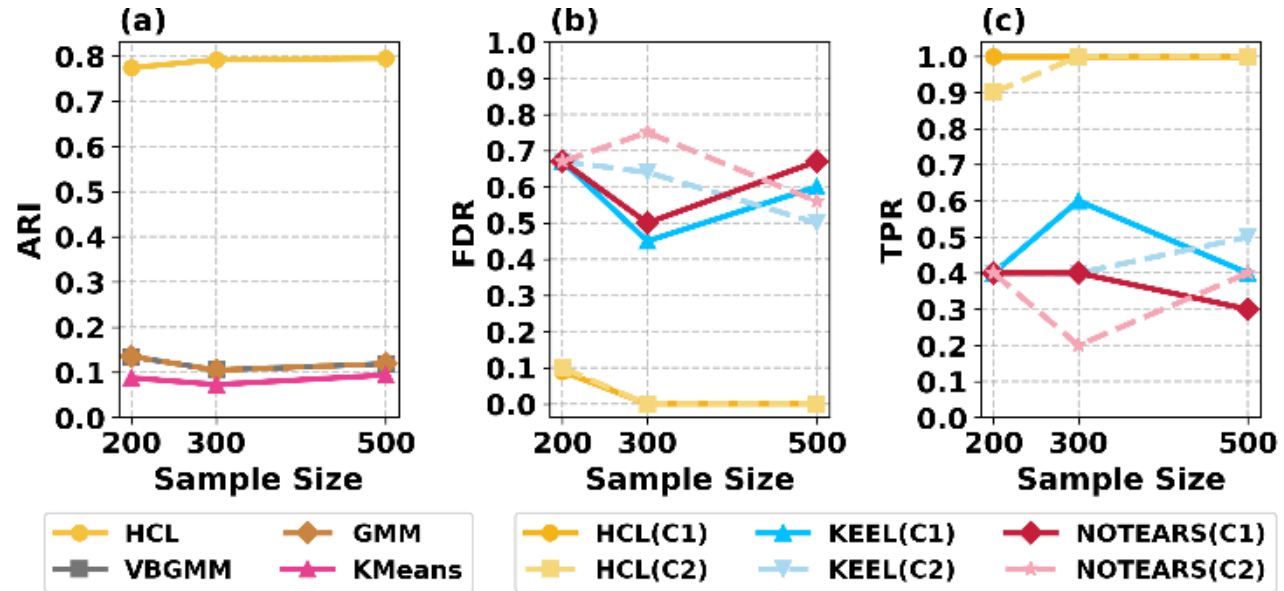

Fig. 2. Results of Datasets 1

The results in Fig. 2 show that HCL consistently outperforms conventional clustering methods across all sample sizes. Specifically, HCL automatically and accurately identifies the true number of clusters and achieves an ARI of 0.774 under a limited sample size of 200, further improving to 0.795 as the sample size increases to 500. In contrast, GMM, VBGMM and KMeans require the true number of clusters (set to 2) as input and consistently yield ARI values below 0.14 across all conditions. Although VBGMM can flexibly infer cluster assignments, its performance remains unsatisfactory due to reliance on purely statistical similarity and inability to account for causal semantics. These results demonstrate the importance of causal structure in enhancing interpretability and cluster separability, particularly when statistical signals are weak. HCL addresses this by leveraging cluster-specific structure learning, which enhance clustering boundaries that are otherwise blurred when causal heterogeneity is ignored. While KEEL and NOTEARS are evaluated per cluster, both assume a globally homogeneous causal mechanism and thus fail to adapt to class-specific causal structures. For instance, when sample size is 200, KEEL exhibits high FDR of 0.67 and low TPR of 0.4 across both classes. NOTEARS performs even worse in terms of both FDR and TPR, and suffers from unstable performance due to sensitivity to noisy partitioning. These findings reveal the limitations of methods that presume causal homogeneity—they struggle to capture structural heterogeneity and generalize poorly to scenarios involving structure shifts. In contrast, HCL demonstrates substantial gains by adaptively modeling heterogeneous causal structures. Even with just 200 samples per class, it achieves TPR values of 1.0 and 0.9 and maintains low FDRs of 0.09 and 0.10 across classes. As the sample size increases, HCL further improves, eventually achieving perfect structure recovery (TPR = 1.0, FDR = 0.0) in both clusters.

These results illustrate that HCL enhances clustering performance by identifying latent structural heterogeneity beyond surface-level features. Its iterative clustering strategy not only enables accurate determination of the number of clusters, but also improves structure learning by iteratively aligning cluster membership with distinct causal modes. Moreover, the integration of equivalent latent representations enables structure-aware clustering, while alternating updates between clustering and DAG estimation promote coherent structural disentanglement. The causality-aware subgraph penalty effectively suppresses overfitting and mitigates noise-induced spurious edges. Additionally, the recomputation of causal mechanisms after clustering facilitates the correction of confounding-induced spurious edges.

Baseline methods, which ignore causal heterogeneity or rely on fixed causal graphs, are inherently limited under heterogeneous data. In contrast, HCL's architecture is uniquely equipped to disentangle underlying mechanisms and robustly recovering accurate structures, even in data-scarce scenarios.

2) **Results of Datasets 2**

As shown in Fig. 3, the proposed HCL method consistently achieves superior clustering performance with high ARI across all imbalance settings, maintaining scores above 0.79 even under the severely imbalanced scenario. In contrast, conventional clustering methods exhibit a dramatic performance drop, with ARI approaching zero or even negative in severely imbalanced settings. This highlights their vulnerability to sample distribution skewness.

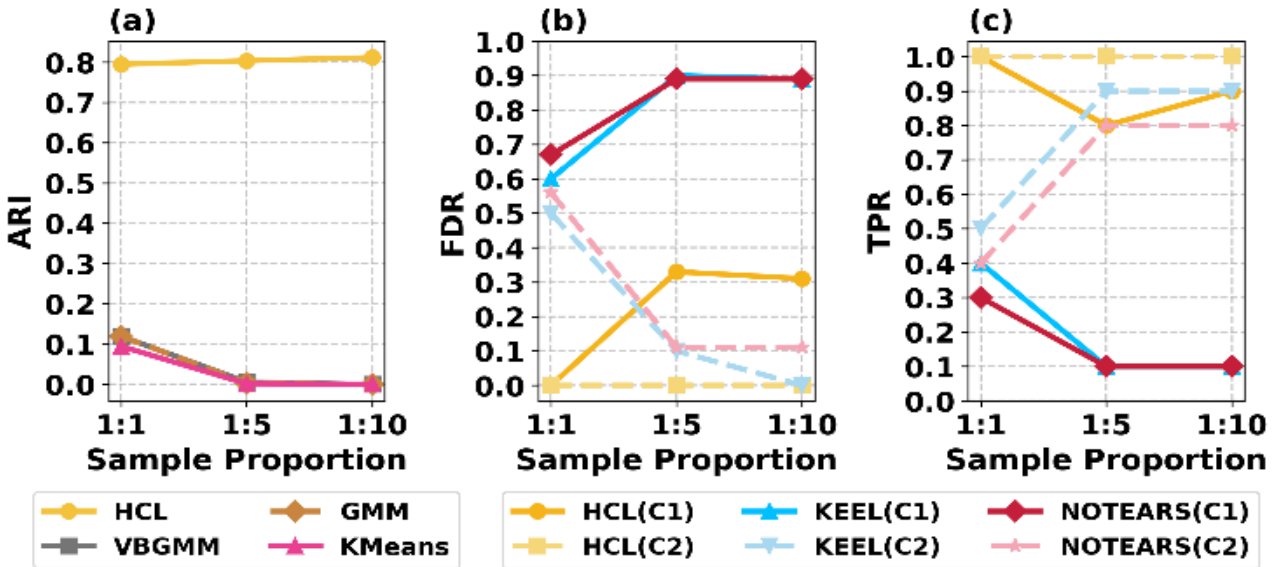

Fig. 3. Results of Datasets 2

The results further reveal the limitations of baseline causal structure learning methods in handling class imbalance. KEEL and NOTEARS primarily capture the structure of the majority class, while the structural patterns of the minority class are increasingly obscured as the imbalance worsens. Consequently, their performance becomes increasingly biased, and the discovery of minority-specific causal mechanisms becomes progressively more difficult with greater class imbalance. In contrast, HCL maintains low FDRs ($<$0.33) and achieves near-perfect TPRs across all proportions, effectively preserving structure estimation quality in both majority and minority classes. These results underscore HCL's robustness in imbalanced scenarios, enabled by its structure-aware clustering and per-class structure learning, which together mitigate bias toward dominant groups and ensure accurate causal identification.

3) **Results of Datasets 3**

As shown in Fig. 4, HCL consistently outperforms baseline methods across all conditions. In terms of clustering performance, HCL maintains highest ARI regardless of structural disparity. Notably, the ARI increases to 0.88 when the edge ratio reaches 20:10, indicating that HCL effectively utilizes structural signals to refine clustering.

For structure learning, baseline methods become increasingly unreliable as structural heterogeneity intensifies. Under the 20:10 edge setting, NOTEARS exhibits an FDR of 0.933 in one cluster, with KEEL showing similarly inflated FDR. Simultaneously, their TPRs degrade markedly—dropping to as low as 0.1—highlighting their inability to distinguish class-specific causal patterns as structural differences grow. These trends indicate that conventional methods struggle to generalize beyond settings with low heterogeneity or fixed structures, limiting their applicability in scenarios with multiple or evolving causal mechanisms.

In contrast, HCL consistently achieves near-zero FDR and perfect TPR across all edge ratios. This robustness illustrates that HCL can accurately recover heterogeneous causal structures regardless of structural disparity or graph degree. Its iterative structure-aligned clustering enables it not only to handle strongly heterogeneous settings but also to uncover subtle differences between groups with similar surface representations. These results validate that HCL is not only generalizable to novel or highly diverse scenarios but is also well-suited for discovering fine-grained structural distinctions often missed by existing approaches.

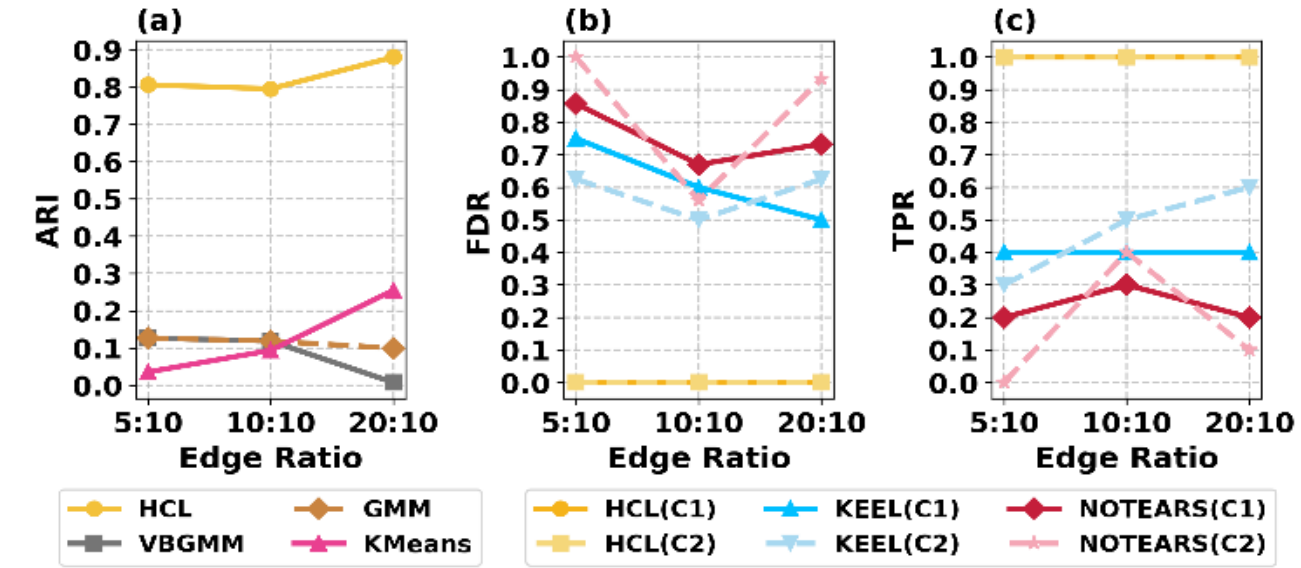

Fig. 4. Results of Datasets 3

4) **Results of Datasets 4**

As shown in Fig. 5, all methods exhibit improved clustering performance as graph complexity increases, which may be attributed to stronger feature interactions and the emergence of richer sample-level representations in denser graphs. In real-world systems characterized by high coupling among variables, such complexity often amplifies latent structural signals, leading to improved clustering for conventional methods. However, this improvement is not uniform. Baseline methods still show marginal or unstable gains, and their performance remains poor under sparse settings (ARI $<$ 0.2), indicating that their reliance on statistical proximity is insufficient to robustly capture sample heterogeneity. In contrast, HCL consistently outperforms all baselines across the full range of complexity, with ARI steadily increasing from 0.689 in sparse graphs to 0.984 in the most complex setting. This result highlights a key distinction in methodology: while existing methods may benefit from incidental representation enrichment in complex graphs, they fail to account for the underlying causal heterogeneity driving sample distinctions. As a result, their clustering performance is sensitive to sparsity, often collapsing in low-complexity scenarios. In contrast, HCL's iterative structure-aligned clustering enables it to extract and utilize the generative mechanisms responsible for sample variation. By disentangling confounding and aligning latent structure with cluster identity, HCL achieves more consistent and accurate clustering, demonstrating the critical role of mechanism-level inference in resolving heterogeneity.

In terms of structure learning, the results further validate HCL's robustness across increasing graph complexity. While baseline methods produce high FDRs (up to 0.8) and maintain low TPRs (≤ 0.4) even as more edges are introduced, HCL consistently improves structure recovery. Under the most complex condition, HCL maintains moderate FDRs (≤ 0.27) and achieves TPRs as high as 0.77, indicating that it effectively utilizes richer structural cues for precise edge identification without overfitting. These findings collectively confirm that HCL is both scalable and resilient across a wide spectrum of structural complexities. Its causal mechanism–driven clustering framework enables it to adapt not only to sparse and noisy scenarios but also to highly entangled systems, making it a robust solution for structure discovery in complex environments.

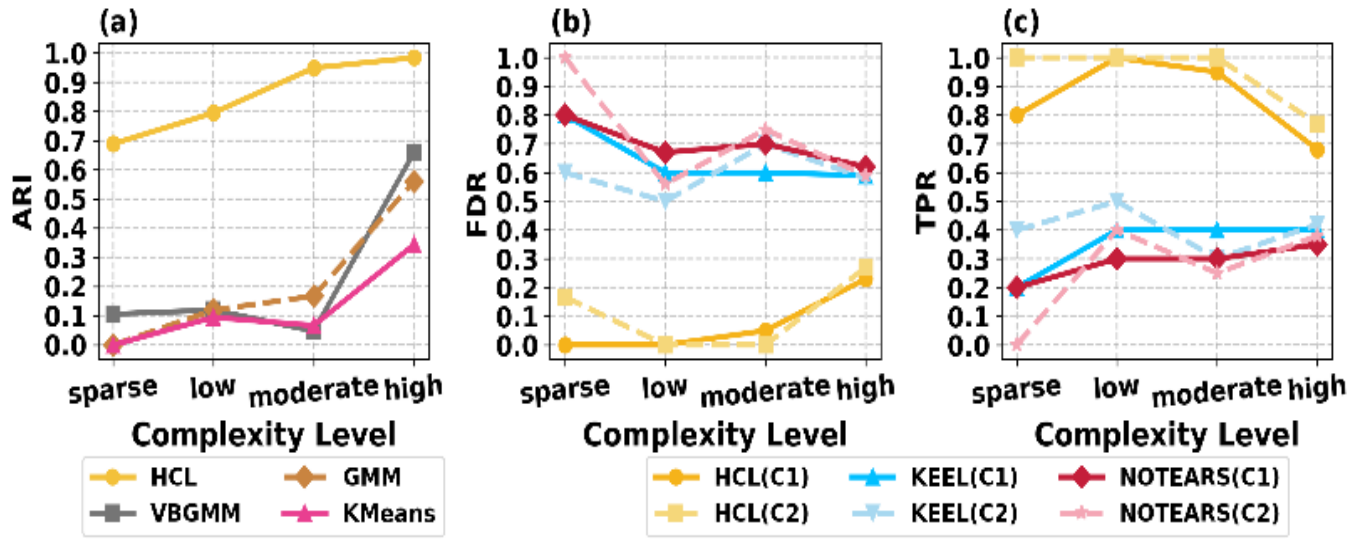

Fig. 5. Results of Datasets 4

5) **Results of Datasets 5**

As shown in Fig. 6, HCL achieves consistently superior performance across all settings. Under the single-cluster condition, HCL achieves perfect clustering and structure recovery (ARI=1.0, TPR=1.0, FDR=0.0909), demonstrating that its iterative refinement process not only avoids over-segmentation but also enhances structure learning by effectively correcting for confounding. As the number of true clusters increases, HCL retains high accuracy through adaptive inference. For example, under the 7-cluster setting, HCL yields an ARI of 0.904, average TPR of 0.9, and average FDR of 0.0875 across clusters, demonstrating strong robustness to structural complexity and heterogeneity. In contrast, VBGMM and DP suffer from severe performance degradation as the number of clusters increases. VBGMM lacks structural awareness and relies solely on statistical proximity, which leads to unstable ARI scores (ARI = 0.173 at 5 clusters). DP, while capable of automatically inferring the number of clusters, tends to overestimate it, resulting in low clustering quality (ARI = 0.095 at 5 clusters, 0.117 at 7).

KEEL and NOTEARS, although equipped with global structure learning capabilities, fail to generalize in the presence of multiple causal patterns. Both show poor performance across clusters with elevated FDRs (often above 0.8) and degraded TPRs (often below 0.2) as cluster number increases, indicating their inability to capture context-specific structures, particularly in highly heterogenous scenarios. In contrast, HCL achieves consistently high structure discovery performance across all clusters, even as the number of causal patterns increases to 7. Specifically, HCL maintains low FDR and high TPR in each individual cluster. This suggests that HCL effectively disentangles cluster-specific causal structures and maintains inference fidelity in the presence of strong heterogeneity. Such robustness highlights HCL's capacity to capture fine-grained causal heterogeneity in complex, multi-mechanism data settings. Even under scenarios with pronounced diversity in causal patterns, HCL accurately recovers cluster assignments and each cluster's unique causal structures. Moreover, these results underscore HCL's potential for real-world applications where data exhibit intricate, latent heterogeneity where accurately resolving diverse causal patterns is essential for interpretability and generalizability.

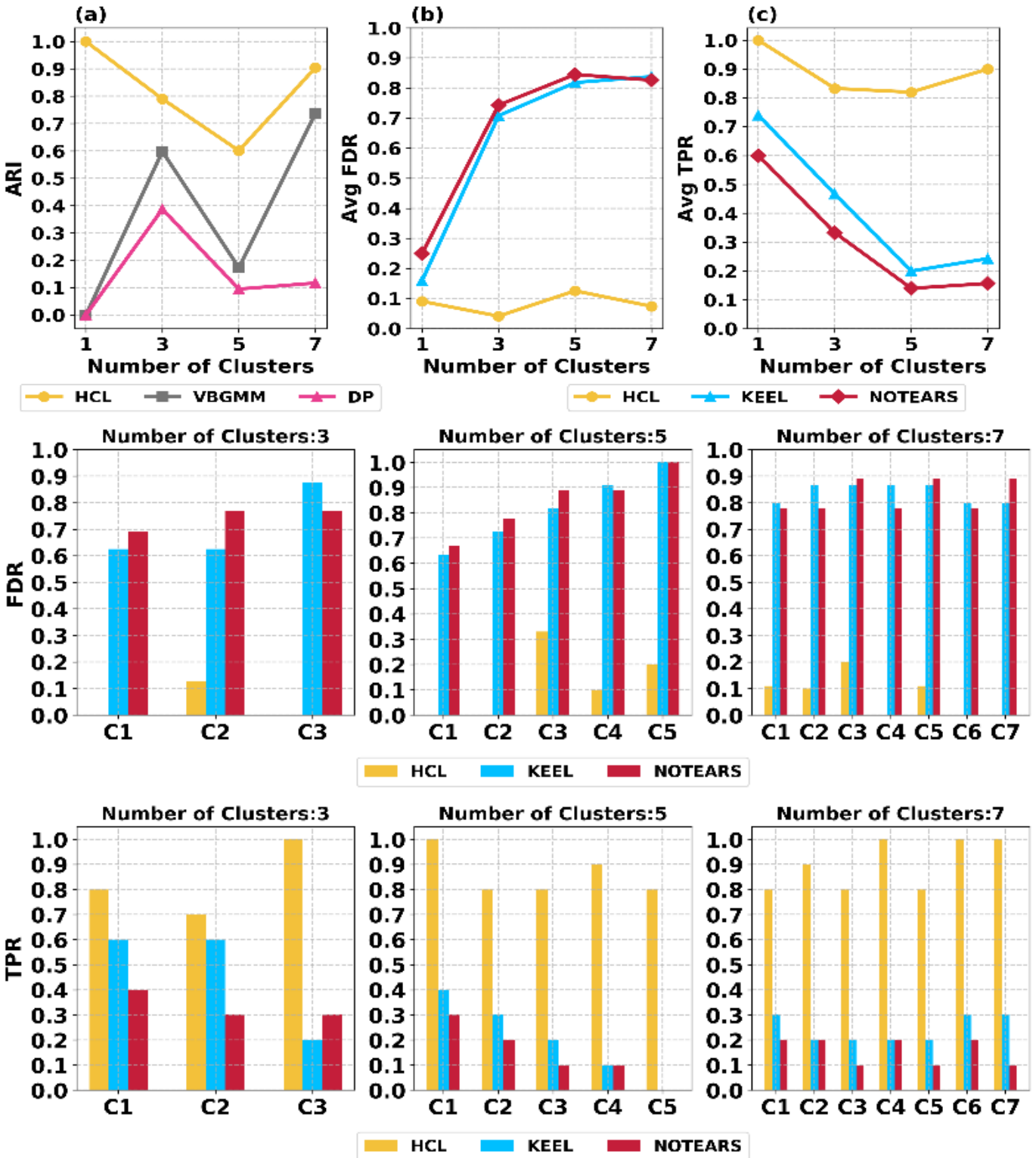

Fig. 6. Results of Datasets 5

*D. Real-World Application*

In this study, we employ HCL to the widely-used protein signaling dataset [47]. This dataset comprises multiparametric flow cytometry measurements across different experimental conditions, including perturbations: CD3+CD28+U0126, CD3+CD28+G06976, and PMA. Perturbation involving CD3+CD28 stimulation activates T cell signaling, and each introduces a different compound targeting specific components. U0126 is a selective inhibitor that blocks MEK activation. G06976 inhibits PKC activity. In contrast, PMA activates PKC. These perturbations represent mechanistically diverse interventions, providing a stringent test for evaluating causal heterogeneity. We select three subsets of the data corresponding to these perturbations. Each subset represents a distinct cellular response to external stimuli, thus inducing variability in the underlying causal mechanisms. These subsets contain measurements of 11 phosphorylated protein and phospholipid variables. We combine these three subsets and use their known perturbation labels as ground truth to evaluate clustering performance.

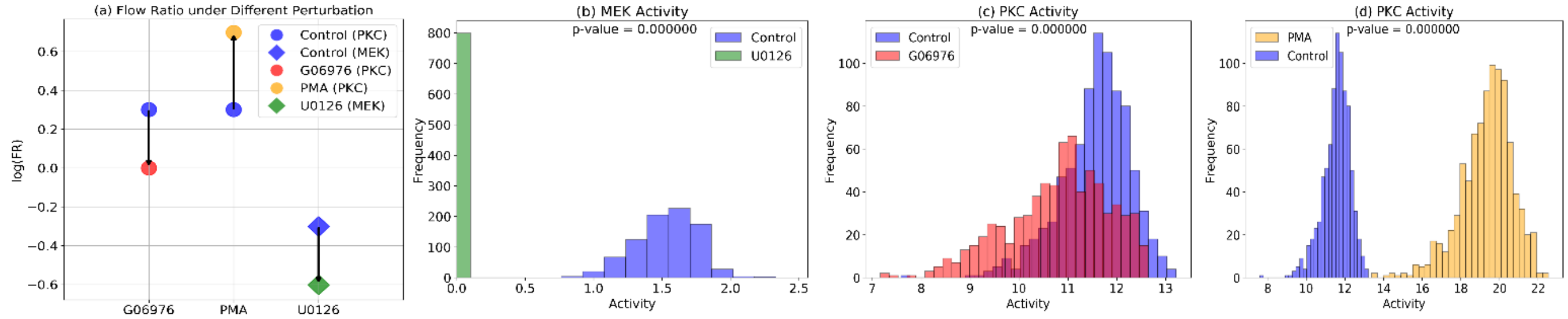

Fig. 7. Interpretability of HCL in discovering distinct cluster-specific causal patterns

To further validate the interpretability of the discovered causal patterns, we utilize the HCL estimated causal graphs and computed the overall downstream influence of each node as a proxy for activity of the corresponding protein. Specifically, for each node, we aggregate its inferred causal effects on all downstream targets across the estimated graph, thereby quantifying its functional impact within the signaling cascade. The resulting node activity scores are compared at the single-cell level across perturbation and control groups. Statistical comparisons are performed using the Wilcoxon rank-sum test to assess differences in activity distributions.

In addition, we introduce the concept of a flow ratio (FR), defined as the ratio of out-degree to in-degree for each node, to assess its role in signal propagation within the causal graph. Nodes with high FR values are presumed to serve as major signal transmitters, while those with low values function more as signal receivers. Using the inferred causal graphs, we quantify the FR under different perturbation conditions.

As shown in Supplementary Table 1, HCL achieves a significantly higher ARI (0.903) compared to alternative methods such as DP (0.496). Notably, DP is found to over-segment the data into 38 clusters, while HCL correctly recovers the ground-truth clusters, reflecting its ability to distinguish mechanistically distinct cellular states. HCL not only attains superior clustering performance but also preserves meaningful cluster granularity. This result validates HCL's ability to accurately recover latent causal modules from complex biological data. Crucially, the strength of HCL lies not only in clustering accuracy but in its inherent interpretability enabled by the explicit modeling of heterogenous causal graphs. Beyond clustering, HCL reveals interpretable, cluster-specific causal structures aligned with known biochemical mechanisms. As illustrated in Fig. 7, HCL detects significant perturbation-specific changes in key signaling proteins. Specifically, the results reveal that MEK activity in the first perturbation group (U0126) is significantly suppressed, which HCL successfully captures as a mechanistic alteration in the causal graph. It is validated by both the HCL-learned heterogenous structure and a highly significant distributional shift (p-value $< 1e\text{-}6$) compared to the control, consistent with the compound's known effect of perturbation. Similarly, Fig. 6 shows that HCL identifies a marked shift in PKC activity under G06976 and PMA perturbations. PKC activity in the second group (G06976) is downregulated, and in the third group (PMA) is markedly upregulated relative to the control—all in concordance with biological inhibitory and activating effects of the respective compounds. These findings underscore the accuracy and biological relevance of the estimated causal structures in capturing perturbation-specific functional changes. Moreover, the FR values highlight the shifts in causal influence under different perturbations. Notably, changes in FR for key signaling molecules MEK and PKC across the three groups are consistent with the expected directionality of perturbation effects. These results demonstrate that the inferred causal structures preserve functionally meaningful directional flow patterns, effectively capturing the heterogeneity of underlying causal mechanisms.

Owing to its unique iterative strategy, HCL does not merely cluster samples based on superficial statistical correlations but faithfully recovers the underlying causal mechanisms specific to each biological condition and actively leverages causal structure variation to uncover biologically meaningful perturbation effects. These results underscore HCL's potential as a powerful tool for mechanism-aware phenotyping, perturbation effect prediction, and causal pattern discovery in complex and heterogeneous biological systems.

## IV. Conclusion

We present **HCL**, a novel unsupervised framework that unifies interpretable clustering and heterogeneous causal structure learning from purely observational data. By relaxing classical assumption of causal homogeneity, HCL enables the discovery of both universal and cluster-specific causal structures without requiring time ordering, environment annotations, interventions or prior knowledge of cluster numbers and structures. Through an equivalent latent representation and a bi-directional iterative strategy, our method adaptively refines clusters, structures, and confounding effects in a mutually reinforcing manner. Theoretically, we establish the identifiability of structural heterogeneity under mild assumptions, and empirically, we demonstrate that HCL outperforms existing methods in both causal structure recovery and interpretable clustering. Application to single-cell protein perturbation data further validates the biological plausibility of the learned structures, highlighting HCL's potential for mechanism-aware phenotyping and structure-guided discovery in complex systems. This work opens new avenues for unsupervised, confounding-resilient causal structure learning in heterogeneous environments. It paves the way toward more interpretable and generalizable models of real-world applications involving complex, heterogeneous systems, such as biology, precision medicine, and social sciences, where unknown mixed populations and context-specific causal mechanisms are ubiquitous.

## APPENDIX

### A. The proof of Proposition 1

First, we compare the two models corresponding to Fig. 1(a) and Fig. 1(b). Then, we proceed to compare the models in Fig. 1(a) and Fig. 1(d). Fig. 1(a) illustrates a homogenous Markov model, where the exogenous latent variables $U_1$ and $U_2$ are independent, $X_1$ and $X_2$ are observed variables, and $X_1$ is a parent of $X_2$. In contrast, Fig. 1(b) depicts a non-Markov model, in which no causal relation exists between $X_1$and $X_2$; instead, their correlation is induced by the latent confounder $U$. Fig. 1(c) further demonstrates that such latent confounding can give rise to a spurious association $X_1 \to X_2$. Fig. 1(d) illustrates a heterogeneous model regulated by the latent modifier $U$. In both cases, they share the same $S$ faithful to $X_1 \to X_2$.

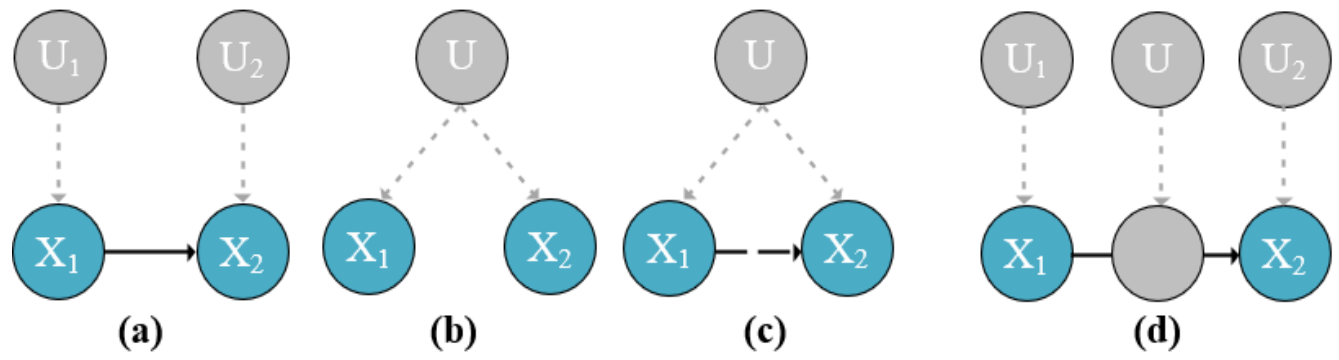

Fig. 1 Illustrations of different global structures.

Denote the mapping function from the value of parents to that of $X$ as $f_X: \Omega_X \to \Omega_Y$. Denote the empirical probability distribution function as $\tilde{p}$. Denote the mapping interval of the child node $X$ when its endogenous parent nodes $pa(X)$ are observed as $T_{X|pa(X)}$:

$$T_{X|pa(X)}(x) = \begin{cases} U, u \in U \leftrightarrow x := \mathrm{f}_X\left(u, pa(X)\right), \text{ x is continuous;} \\ U_\alpha, P\left(u \mid x \in U_\alpha\right) = 1-\alpha \leftarrow x := \mathrm{f}_X\left(u, pa(X)\right), \\ u \in U_\alpha \to x := \mathrm{f}_X\left(u, pa(X)\right), \text{ x is discrete.} \end{cases}$$

Considering model (a), according to decomposition property of the Markov model:

$$p_a\left(x_1, x_2\right) = \tilde{p}\left(x_1\right)\tilde{p}\left(x_2 \mid x_1\right).$$

Considering model (b), due to the presence of confounding, the decomposition property of the Markov model is no longer applicable:

$$p_b\left(x_1, x_2\right) \neq \tilde{p}\left(x_1\right)\tilde{p}\left(x_2\right).$$

The decomposition property is as follows:

$$p(x) = \prod_x \tilde{p}\left(x|y_X\right).$$

Here, $y_X$ represents the true parent nodes $pa(X)$ along with all variables with the same topological order as $X$. The latent confounder causes the estimation of $pa(X)$ to be $y_X$. Therefore, estimating the joint distribution from observational data, we have:

$$p_b\left(x_1, x_2\right) = \tilde{p}\left(x_1\right)\tilde{p}\left(x_2 \mid x_1\right)$$

Thus, model (a) and (b) yield the same joint distribution of $X$. The true structure is unidentifiable based on $p(X)$. Proposition (1) is proved.

Further, consider the two model (a) and (b) in the mapped probability space. For model (a):

$$\begin{aligned} p_a\left(\hat{x} \mid x, S\right) &= \int_{\hat{u}_1 \in T_{X_1}(x_1)} p(\hat{u}_1)\,\mathrm{d}\hat{u}_1 \int_{\hat{u}_2 \in T_{X_2|x_1}(x_2)} p(\hat{u}_2)\,\mathrm{d}\hat{u}_2 \\ &= \int_{u_1 \in T_{X_1}(x_1)} p(u_1)\,\mathrm{d}u_1 \int_{u_2 \in T_{X_2|x_1}(x_2)} p(u_2)\,\mathrm{d}u_2 \\ &= \int_{a \in T_{X_1}(x_1)} \psi(a)\,\mathrm{d}a \int_{b \in T_{X_2|x_1}(x_2)} \psi(b)\,\mathrm{d}b = \tilde{p}(x_1)\tilde{p}(x_2 \mid x_1). \end{aligned}$$

For model (b):

$$\begin{aligned} p_b\left(\hat{x} \mid x, S\right) &= \iint_{\substack{\hat{u}_1 \in \tilde{T}_{X_1}(x_1), \\ \hat{u}_2 \in \tilde{T}_{X_2|x_1}(x_2)}} p(\hat{u}_1)p(\hat{u}_2)\,\mathrm{d}\hat{u}_1\,\mathrm{d}\hat{u}_2 \\ &= \iint_{\substack{\hat{u}_1 \in \tilde{T}_{X_1}\left(f_{X_1}(u)\right), \\ \hat{u}_2 \in \tilde{T}_{X_2|x_1}\left(f_{X_2}(u)\right)}} p(\hat{u}_1)p(\hat{u}_2)\,\mathrm{d}\hat{u}_1\,\mathrm{d}\hat{u}_2 \\ &= \int_{u \in T_{X1}(x1) \cap T_{X2}(x2)} \psi\left(\hat{u}_1(u)\right)\psi\left(\hat{u}_2(u)\right)\sqrt{\left(\frac{\mathrm{d}\hat{u}_1}{\mathrm{d}u}\right)^2 + \left(\frac{\mathrm{d}\hat{u}_2}{\mathrm{d}u}\right)^2}\,\mathrm{d}u \\ &\neq \int_{u \in T_{X1}(x1) \cap T_{X2}(x2)} \psi(u)\,\mathrm{d}u\,; \end{aligned}$$

$$\int_{u \in T_{X1}(x1) \cap T_{X2}(x2)} \psi(u)\,\mathrm{d}u = \tilde{p}\left(x_1\right)\tilde{p}\left(x_2 \mid x_1\right).$$

Thus:

$$p_b\left(\hat{x} \mid x, S\right) \neq \tilde{p}\left(x_1\right)\tilde{p}\left(x_2 \mid x_1\right).$$

According to (1)-(2), we have:

$$z_a = \left(\varphi\left(x_1, \hat{x}_{1a}, \boldsymbol{\theta}\right), \varphi\left(x_2, \hat{x}_{2a}, \boldsymbol{\theta}\right)\right),$$

$$z_b = \left(\varphi\left(x_1, \hat{x}_{1b}, \boldsymbol{\theta}\right), \varphi\left(x_2, \hat{x}_{2b}, \boldsymbol{\theta}\right)\right).$$

We define:

$$\mathrm{G}: \mathbb{R}^2 \to \mathbb{R}^2.\ \mathrm{G}\left(x_1, x_2\right) = \left(\mathrm{g}\left(x_1\right), \mathrm{g}\left(x_2\right)\right).$$

It is straightforward that if g is injective, then G is also injective. Conditioned on $x, S$ and given the parameter $\boldsymbol{\theta}$, since $\varphi$ is an injective function of $\hat{x}$, it follows that $z$ is an injective function of $\hat{x}$. Under this condition,

$$p_a\left(\hat{x} \mid x, S\right) \neq p_b\left(\hat{x} \mid x, S\right).$$

Therefore:

$$p(z_a \mid x, S) \neq p(z_b \mid x, S).$$

Proposition (2) is proved.

Considering model (d), the following property holds:

$$p_d\left(x_1, x_2\right) = \tilde{p}\left(x_1\right)\tilde{p}\left(x_2 \mid x_1\right).$$

Thus, model (a) and (d) yield the same joint distribution of $X$. The genuine structure heterogeneity is unidentifiable based on $p(X)$. Proposition (1) is proved.

Further, consider the model (d) in the mapped probability space. Suppose that from the entire sample set we obtain an estimator $\tilde{T}_{X|pa(X)}$. For model (d), there exists $X = x$, such that $T_{X|pa(X)}(x) \neq \tilde{T}_{X|pa(X)}(x)$. Thus, there exists:

$$\begin{aligned} p_d\left(\hat{x} \mid x, S\right) &= \int_{\hat{u}_1 \in T_{X_1}(x_1)} p(\hat{u}_1)\,\mathrm{d}\hat{u}_1 \int_{\hat{u}_2 \in \tilde{T}_{X_2|x_1}(x_2)} p(\hat{u}_2)\,\mathrm{d}\hat{u}_2 \\ &\neq \int_{\hat{u}_1 \in T_{X_1}(x_1)} p(\hat{u}_1)\,\mathrm{d}\hat{u}_1 \int_{\hat{u}_2 \in T_{X_2|x_1}(x_2)} p(\hat{u}_2)\,\mathrm{d}\hat{u}_2. \end{aligned}$$

This implies that:

$$p_d\left(\hat{x} \mid x, S\right) \neq \tilde{p}\left(x_1\right)\tilde{p}\left(x_2 \mid x_1\right).$$

Namely,

$$p_d\left(\hat{x}\mid x,S\right)\neq p_a\left(\hat{x}\mid x,S\right).$$

By similar reasoning,

$$p(z_a\mid x,S)\neq p(z_d\mid x,S)\ .$$

Proposition (2) is proved.

### *B. The proof of Proposition 3*

Suppose $\exists\epsilon>0$, such that for $O$ index set $I$ and some $i,j,$ where $i\in I,j\in I\backslash\{i\}$:

$$\lim_{N\to\infty}\mathrm{P}(C^Z=i\mid C^Z=j)\geq\epsilon.$$

Then the distribution of $Z$ assigned to cluster $i$, denoted $\hat{P}_i^Z$, will contain a nonzero portion of $P_j^Z$, the true distribution under $O^j$. Since the conditional expectations of latent representations are distinguishable across regimes as proved in Section A, we have:

$$\|\mu_i^Z-\mu_j^Z\|_2\geq\delta>0,\forall i\neq j,$$

and consequently $D_{\mathrm{KL}}(P_j^Z||P_i^Z)\geq\eta$. By the convexity of KL divergence, it follows that:

$$D_{\mathrm{KL}}(\hat{P}_i^Z\parallel P_j^Z)>0\ .$$

On the other hand, the Bayesian Gaussian mixture model is consistent [48], implying:

$$D_{\mathrm{KL}}(\hat{P}_i^Z\parallel P_j^Z)\to 0.$$

This leads to contradiction unless misalignment probability vanishes. Therefore,

$$\lim_{N\to\infty}P(C^Z=i\mid C^Z=j)=0.$$

Thus, incorrect assignments to mutually exclusive groups are eventually precluded.

Further, consider any two clusters $c_1$ and $c_2$, with estimated structures $\hat{G}^{c_1},\hat{G}^{c_2}$, respectively.

If both clusters are primarily composed of samples from the same generating mechanism $G^k$, then:

$$d_{\mathrm{SHD}}\left(\hat{G}^{c_1},\hat{G}^{c_2}\right)\leq d_{\mathrm{SHD}}\left(\hat{G}^{c_1},G^k\right)+d_{\mathrm{SHD}}\left(\hat{G}^{c_2},G^k\right)\xrightarrow{P}0.$$

Hence, these clusters will eventually be correctly merged under the refinement rule $d_{\mathrm{SHD}}<\gamma$, with $\gamma\to 0$.

If the two clusters are primarily composed of samples from distinct mechanisms $G^k\neq G^j$, then under (A1), it can be implied:

$$d_{\mathrm{SHD}}\left(\hat{G}^{c_1},\hat{G}^{c_2}\right)\geq d_{\mathrm{SHD}}\left(G^k,G^j\right)-d_{\mathrm{SHD}}\left(\hat{G}^{c_1},G^k\right)-d_{\mathrm{SHD}}\left(\hat{G}^{c_2},G^j\right)\xrightarrow{P}\delta_{\min}^{\mathrm{SHD}}+\varepsilon.$$

Therefore, with high probability:

$$d_{\mathrm{SHD}}\left(\hat{G}^{c_1},\hat{G}^{c_2}\right)\geq\delta_{\min}^{\mathrm{SHD}}>\gamma.$$

Hence, incorrect merges between structurally distinct clusters are eventually precluded.

### *C. Details of Real-World Application*

**Table S1.** Clustering results on the protein signaling datasets under different perturbations

| **Methods** | GMM | KMeans | VBGMM | DP | HCL |
|---|---|---|---|---|---|
| **ARI** | 0.467 | 0.011 | 0.441 | 0.496 | 0.903 |

## References

[1] A. Zanga, E. Ozkirimli, and F. Stella, “A Survey on Causal Discovery: Theory and Practice,” *Int. J. Approx. Reason.*, vol. 151, pp. 101–129, Dec. 2022.

[2] J. Pearl and D. Mackenzie, The Book of Why: The New Science of Cause and Effect, New York, NY, USA: Basic Books, 2018.

[3] C. M. Gilligan-Lee, C. Hart, J. Richens, and S. Johri, “Leveraging directed causal discovery to detect latent common causes in cause-effect pairs,” *IEEE Trans. Neural Netw. Learn. Syst.*, vol. 35, no. 4, pp. 4938–4947, Apr. 2024.

[4] Fu, X., Mo, S., and Buendia, A., et al., “A foundation model of transcription across human cell types,” *Nature*, vol. 637, no. 8047, pp. 965–973, Jan. 2025.

[5] Feuerriegel, S., Frauen, D., Melnychuk, V., et al., “Causal machine learning for predicting treatment outcomes,” *Nat Med*, vol. 30, no. 4, pp. 958–968, Apr. 2024.

[6] You, W., Zhang, Y., and Lee, C. C., “The dynamic impact of economic growth and economic complexity on CO2 emissions: An advanced panel data estimation,” *Economic Analysis and Policy*, vol. *73*, pp. 112-128, Mar. 2022.

[7] J. Peters, D. Janzing and B. Schölkopf, Elements of Causal Inference: Foundations and Learning Algorithms, Cambridge, MA, USA: MIT Press, 2017.

[8] J. Pearl, Causality: Models Reasoning and Inference, Cambridge, U.K.: Cambridge Univ. Press, 2011.

[9] A. Tejada-Lapuerta, P. Bertin, S. Bauer, et al., “Causal machine learning for single-cell genomics,” *Nat Genet*, vol. 57, no. 4, pp. 797–808, Apr. 2025.

[10] C. Glymour, K. Zhang and P. Spirtes, "Review of causal discovery methods based on graphical models," *Front. Genet.*, vol. 10, Jun. 2019.

[11] Y. Wei, X. Li, L. Lin, et al., “Causal discovery on discrete data via weighted normalized Wasserstein distance,” *IEEE Trans. Neural Netw. Learn. Syst.*, vol. 35, no. 4, pp. 4911–4923, Apr. 2024.

[12] Z. Cai, D. Xi, X. Zhu, et al., “Causal discoveries for high dimensional mixed data,” *Stat. Med.*, vol. 41, no. 24, pp. 4924–4940, Oct. 2022.

[13] A. R. Nogueira, A. Pugnana, S. Ruggieri, et al., "Methods and tools for causal discovery and causal inference", *Wiley Interdiscipl. Rev.: Data Mining Knowl. Discov.*, vol. 12, no. 2, e1449, Apr. 2022.

[14] B. Huang, K. Zhang and Y. Lin, "Generalized score functions for causal discovery", *Proc. 24th ACM SIGKDD*, pp. 1551-1560, Jul. 2018.

[15] Y. Li, R. Xia, C. Liu, et al., “A Hybrid Causal Structure Learning Algorithm for Mixed-Type Data,” *Proc. AAAI Conf. Artif. Intell.*, vol. 36, no. 7, pp. 1435-1443, Jun. 2022.

[16] X. Zheng, C. Dan, B. Aragam, et al., "Learning sparse nonparametric DAGs", *Proc. Int. Conf. Artif. Intell. Stat.*, pp. 3414-3425, 2020.

[17] W. Li, W. Zhang, Q. Zhang, et al., "Weakly Supervised Causal Discovery Based on Fuzzy Knowledge and Complex Data Complementarity," *IEEE Transactions on Fuzzy Systems*, vol. 32, no. 12, pp. 7002-7014, Dec. 2024.

[18] Eric V. Strobl. "Causal discovery with a mixture of dags," *Machine Learning*, vol. 112, no. 11, pp. 4201–4225, Nov. 2023.

[19] B. Huang, K. Zhang, M. Gong, et al., “Causal Discovery and Forecasting in Nonstationary Environments with State-Space Models,” *Proc Mach Learn Res,* vol. 97, pp. 2901-2910, Jun. 2019.

[20] B. Huang, K. Zhang, J. Zhang, et al., “Causal discovery from heterogeneous/nonstationary data,” *Journal of Machine Learning Research*, vol. 21, no. 89, pp. 1-53, May. 2020.

[21] K. Zhang, B. Huang, J. Zhang, et al., “Causal discovery from nonstationary/heterogeneous data: Skeleton estimation and orientation determination,” *IJCAI,* vol. 2017, p. 1347, Aug. 2017.

[22] Varıcı, B., Katz, D., Wei, D., et al., “Interventional causal discovery in a mixture of DAGs,” *Advances in Neural Information Processing Systems*, vol. *37*, pp. 86574-86601, 2024.

[23] Löwe, S., Madras, D., Zemel, R., et al., “Amortized causal discovery: Learning to infer causal graphs from time-series data,”. *Conference on Causal Learning and Reasoning, PMLR,* vol.177, pp. 509-525, Jun. 2022.

[24] Zheng X, Sun X, Chen W, et al., “Causally invariant predictor with shift-robustness,” *stat*, vol.1050, p. 5, Jul. 2021.

[25] Yang, D., He, X., Wang, J., et al., "Federated causality learning with explainable adaptive optimization," *Proceedings of the AAAI Conference on Artificial Intelligence,* vol. 38, no. 15, pp. 16308-16315, Mar. 2024.

[26] F. Cao, Y. Wang, K. Yu, et al., "Causal Discovery from Unknown Interventional Datasets Over Overlapping Variable Sets," *IEEE Transactions on Knowledge and Data Engineering*, vol. 36, no. 12, pp. 7725-7742, Dec. 2024.

[27] Markham, A., Das, R., and Grosse-Wentrup, M., "A distance covariance-based kernel for nonlinear causal clustering in heterogeneous populations," *Conference on Causal Learning and Reasoning, PMLR,* vol. *177,* pp. 542-558, Jun. 2022.

[28] Jabbari, F., and Cooper, G. F., "An instance-specific algorithm for learning the structure of causal Bayesian networks containing latent variables," *Proceedings of the 2020 SIAM International Conference on Data Mining,* pp. 433-441, 2020.
[29] B. Huang, K. Zhang, P. Xie, et al., "Specific and shared causal relation modeling and mechanism-based clustering," *Proc. Advances in Neural Information Processing Systems,* vol.32, no.1211, pp. 13520-13531, Dec. 2019.
[30] K. Zhang and M. R. Glymour, "Unmixing for causal inference: Thoughts on Mccaffrey and Danks," *The British Journal for the Philosophy of Science*, vol. 71, no.4, pp. 1319–1330, Dec. 2020.
[31] Kim, K., Kim, J., Wasserman, L., et al., "Hierarchical and density-based causal clustering". *Advances in Neural Information Processing Systems*, vol. *37*, no. 956, pp. 30363-30393, Dec. 2024.
[32] Perry, R., Von Kügelgen, J., and Schölkopf, B., "Causal discovery in heterogeneous environments under the sparse mechanism shift hypothesis," *Advances in Neural Information Processing Systems*, vol. 35, no.792, pp. 10904-10917, Nov. 2022.
[33] Varambally, S., Ma, Y., and Yu, R., "Discovering Mixtures of Structural Causal Models from Time Series Data," *International Conference on Machine Learning,* vol.41, no.2011, pp. 49171-49202, Jul. 2024.
[34] Günther, W., Popescu, O. I., Rabel, M., et al., "Causal discovery with endogenous context variables," *Advances in Neural Information Processing Systems*, vol. 37, no. 1143, pp. 36243-36284, Jun 2025.
[35] Mooij, J. M., Magliacane, S., and Claassen, T., "Joint causal inference from multiple contexts," *Journal of machine learning research*, vol. 21, no. 99, pp. 1-108, Mar. 2020.
[36] Yin, M., Wang, Y., and Blei, D. M., "Optimization-based causal estimation from heterogeneous environments," *Journal of Machine Learning Research*, vol. *25,* no. 168, pp. 1-44, Apr. 2024.
[37] Karlsson, R., and Krijthe, J. "Detecting hidden confounding in observational data using multiple environments," *Advances in Neural Information Processing Systems*, vol. *36*, no. 1916, pp. 44280-44309, Dec. 2023.
[38] C. Liu and K. Kuang, "Causal structure learning for latent intervened non-stationary data," *Proc. International Conference on Machine Learning*, vol. 40, no. 900, pp. 21756-21777, Jul. 2023.
[39] Zhou, F., He, K., and Ni, Y., "Causal discovery with heterogeneous observational data," *Uncertainty in Artificial Intelligence*, *PMLR*, vol. 180, pp. 2383-2393, Aug. 2022.
[40] B. Huang, K. Zhang, M. Gong, et al.,"Causal discovery from multiple data sets with non-identical variable sets," *Proceedings of the AAAI conference on artificial intelligence*, vol. 34, no. 6, pp. 10153-10161, Apr. 2020.
[41] B. Varıcı, D. Katz, D. Wei, et al., "Separability analysis for causal discovery in mixture of DAGs," *Trans. Mach. Learn. Res.*, Jan. 2024.
[42] J. Attia, E. Holiday, C. Oldmeadow, "A proposal for capturing interaction and effect modification using DAGs," *International Journal of Epidemiology*, vol. 51, no. 4, pp. 1047-1053, Aug. 2022.
[43] P. Erdős and A. Rényi, "On random graphs. I," *Publ. math. debrecen*, vol. 6, no. 290-297, p. 18, 1959.
[44] C. M. Bishop, *Pattern Recognition and Machine Learning*. New York, NY, USA: Springer, 2006.
[45] H. Attias, "A variational Bayesian framework for graphical models," *Advances in Neural Information Processing Systems*, vol. 12, pp. 209-215, 2000.
[46] D. M. Blei and M. I. Jordan, "Variational inference for Dirichlet process mixtures," *Bayesian Anal.*, vol. 1, no. 1, pp. 121–143, March. 2006.
[47] K. Sachs, O. Perez, D. Pe'er, et al., "Causal protein-signaling networks derived from multiparameter single-cell data", *Science*, vol. 308, no. 5721, pp. 523-529, Apr. 2005.
[48] Wang, B., Titterington, D. M., "Convergence properties of a general algorithm for calculating variational Bayesian estimates for a normal mixture model," Bayesian Analysis, vol. 1, no. 3, pp.625-650, 2006.